\documentclass[sigconf]{acmart}

\usepackage{graphicx} 
\usepackage{kotex}   
\usepackage{comment}
\usepackage{xspace}
\usepackage{bm}
\usepackage[normalem]{ulem}
\usepackage{multirow}
\usepackage{colortbl}
\usepackage{multicol}
\usepackage{graphics}
\usepackage{pifont}
\usepackage[dvipsnames]{xcolor} 
\usepackage{enumitem}
\usepackage{footmisc}
\usepackage[most]{tcolorbox} 

\tcbset{
  promptstyle/.style={
    enhanced,
    breakable,             
    colback=gray!10,       
    colframe=black!70,     
    left=6pt, right=6pt,
    top=6pt, bottom=6pt,
    boxrule=0.5pt,
    arc=2pt,               
    fonttitle=\bfseries,
    title={#1}
  }
}

\usepackage{longtable}   

\usepackage{color}
\definecolor{gray}{rgb}{0.4,0.4,0.4}
\definecolor{blue}{rgb}{0.0,0.0,0.6}
\definecolor{cyan}{rgb}{0.0,0.6,0.6}
\definecolor{maroon}{rgb}{0.5,0,0}
\definecolor
{darkgreen}{rgb}{0,0.5,0}
\definecolor{lightgray}{rgb}{0.8,0.8,0.8}

\usepackage{xcolor}
\definecolor{codegreen}{rgb}{0,0.6,0}
\definecolor{codegray}{rgb}{0.5,0.5,0.5}
\definecolor{codepurple}{rgb}{0.58,0,0.82}
\definecolor{backcolour}{rgb}{0.95,0.95,0.92}

\newcommand{\cmark}{{\ding{51}}} 
\newcommand{\xmark}{{\ding{55}}} 

\newcommand{\sys}{\textsf{MobiBench}\xspace}

\newcommand{\red}[1]{{\color{red}{#1}}}
\newcommand{\green}[1]{{\color{green}{#1}}}

\usepackage{tikz}
\usetikzlibrary{positioning,arrows,calc,automata}

\usepackage{booktabs,multirow,threeparttable,makecell}
\usepackage{siunitx}
\usepackage{pifont}
\newcommand{\best}[1]{\textbf{#1}}

\title[MobiBench]{MobiBench: Multi-Branch, Modular Benchmark for Mobile GUI Agents}

\author{Youngmin Im$^{1,*}$, Byeongung Jo$^{2,*}$, Jaeyoung Wi$^1$, Seungwoo Baek$^2$,
\\Tae Hoon Min$^2$, Joo Hyung Lee$^2$, Sangeun Oh$^3$, Insik Shin$^{1,4}$, Sunjae Lee$^2$}
\affiliation{
    \institution{$^1$KAIST, S. Korea \quad$^2$Sungkyunkwan University, S. Korea,  \quad$^3$Korea University, S. Korea \quad$^4$Fluiz, S. Korea}
    \country{}
}
\email{ym.im@kaist.ac.kr}
\email{{whale1510,sunjae.lee}@skku.edu}

\renewcommand{\shortauthors}{Y. Im, B. Jo, et al.}

\begin{document}
\begin{abstract}
Mobile GUI Agents---AI agents capable of interacting with mobile applications on behalf of users---have the potential to transform human-computer interaction. However, current evaluation practices for GUI agents face two fundamental limitations. First, they either rely on single-path offline benchmarks or online live benchmarks: Offline benchmarks unfairly penalize valid alternative paths, while online benchmarks lack scalability and reproducibility. Second, existing benchmarks treat agents as monolithic black boxes, overlooking the contributions of individual components and obscuring performance bottlenecks. We present \sys, the first multi-path aware and modular offline benchmarking framework for Mobile GUI Agents that enables high-fidelity, scalable, and reproducible evaluation entirely in offline settings. \sys achieves 94.72\% agreement with human evaluators---on par with carefully engineered online benchmarks---while preserving the scalability and reproducibility of static offline benchmarks. Furthermore, our comprehensive module-level analysis uncovers several key insights and actionable guidelines for designing more capable and cost-efficient mobile agents.

\end{abstract}

\maketitle
\section{Introduction}
Mobile intelligent agents hold the promise of transforming human-computer interaction, alleviating users from tedious and cumbersome smartphone operations. Empowered by Large Foundation Models (LFMs), these agents can now autonomously navigate complex graphical user interfaces (GUIs) on behalf of humans \cite{appagent,mobile-agent-v2,explore-select-derive,uitars,autodroid}. However, as these agents become more sophisticated, evaluating their true capabilities remains a fundamental challenge. Current methodologies \cite{aitw,androidworld,mobile-bench,mobile-bench-v2,llamatouch,androidarena,mobileagentbench,mobile-env,android-lab, motif} struggle to capture the dynamic, non-linear ways people actually use mobile apps, hindering our ability to systematically design and improve these systems.

Existing benchmarks fall into two categories: offline or online evaluation, each with its own limitations. Offline benchmarks \cite{aitw,amex,AITZ,motif,mobile-bench-v2,mobilegpt,metagui} rely on static datasets where each step is paired with a single "correct" action, treating any deviation from this pre-recorded "golden path" as failure. However, mobile interfaces rarely offer only one valid way to accomplish a task. For instance, when booking a flight, one could tap the ‘Search Flights’ button, type a destination directly into the search bar, or select an ‘Explore Trips’ shortcut. Although all options are perfectly valid, single-path datasets would unfairly penalize any decision that does not follow the \textit{"golden"} path. While static datasets are easy to scale and provide a reproducible evaluation environment, their rigidity fails to capture the path-diversity in mobile tasks, leading to systematic underestimation of agent capabilities and unreliable evaluation results.

Online benchmarks \cite{androidworld,llamatouch,android-lab,mobile-bench,mobileagentbench,mobile-env,androidarena} address this limitation by running agents in live environments, where task success is determined by app-state checkpoints rather than fixed action sequences. While this naturally accommodates multiple valid paths, it introduces severe scalability and reproducibility challenges. Crafting reliable checkpoints demands intimate knowledge of each app's internal logic, often requiring code-level instrumentation. More critically, mobile apps evolve continuously: UI layouts change across updates, new features are introduced, and existing workflows are restructured. Each such change can silently invalidate the checkpoints that evaluators have carefully engineered, requiring recurring human effort to diagnose, repair, and re-validate the setup. This continuous maintenance burden severely limits scalability, and the difficulty of reproducing identical app states across evaluation runs undermines reproducibility. As a result, most online benchmarks remain restricted to a handful of simplified applications, far from the complex interfaces real users interact with daily.


Furthermore, existing benchmarks \cite{aitw,mobile-bench,mobileagentbench} treat agents as monolithic black boxes, evaluating only end-to-end performance without distinguishing the contributions of individual components within the agentic system. This coarse-grained evaluation prevents researchers from identifying performance bottlenecks, optimizing specific modules, and leads to unfair comparisons between agents that employ different underlying components.


\begin{table*}[]
\caption{Comparison of \sys to other benchmarks. \textnormal{MobiBench is the first offline benchmarking framework to support multi-path evaluation of mobile GUI agents while preserving the scalability and reproducibility of offline evaluation. In addition, it enables modular assessment of mobile GUI agents, facilitating deeper insight into their true performance.} }
\small
\begin{tabular}{cccccc}
\hline
\textbf{Benchmark} &
  \begin{tabular}[c]{@{}c@{}}\textbf{Modular}\\ \textbf{Assessment}\end{tabular} &
  \begin{tabular}[c]{@{}c@{}}\textbf{Multi-path}\\ \textbf{Evaluation}\end{tabular} &
  \begin{tabular}[c]{@{}c@{}}\textbf{Scalable}\\ \textbf{Dataset}\end{tabular} &
  \begin{tabular}[c]{@{}c@{}}\textbf{Reproducible}\\ \textbf{Results}\end{tabular} &
  \begin{tabular}[c]{@{}c@{}}\textbf{Real world}\\ \textbf{Apps}\end{tabular} \\ \hline
\multicolumn{6}{c}{Offline Benchmarks} \\ \hline
AITW~\cite{aitw}             &\red{\xmark}&\red{\xmark}&\green{\cmark}&\green{\cmark}&\green{\cmark}\\
 MoTiF~\cite{motif}& \red{\xmark}& \red{\xmark}& \green{\cmark}& \green{\cmark}&\green{\cmark}\\
 Meta-Gui~\cite{metagui}& \red{\xmark}& \red{\xmark}& \red{\xmark}& \green{\cmark}&\green{\cmark}\\
Mobile-Bench-v2~\cite{mobile-bench-v2}& \red{\xmark}& \green{\cmark}& \red{\xmark}& \green{\cmark}&\green{\cmark}\\ 
 MobileGPT~\cite{mobilegpt}& \red{\xmark}& \red{\xmark}& \green{\cmark}& \green{\cmark}&\green{\cmark}\\
 DroidTask~\cite{autodroid}& \red{\xmark}& \red{\xmark}& \green{\cmark}& \green{\cmark}&\green{\cmark}\\ \hline
\multicolumn{6}{c}{Online Benchmarks}\\ \hline
 AndroidArena~\cite{androidarena}     & \red{\xmark}& \green{\cmark}& \green{\cmark}& \red{\xmark}&\green{\cmark}\\
LlamaTouch~\cite{llamatouch}       &\red{\xmark}&\green{\cmark}&\red{\xmark}&\red{\xmark}&\green{\cmark}\\
Mobile-Bench~\cite{mobile-bench}& \red{\xmark}& \green{\cmark}& \red{\xmark}& \red{\xmark}&\green{\cmark}\\
MobileAgentBench~\cite{mobileagentbench} &\red{\xmark}&\green{\cmark}&\red{\xmark}&  \red{\xmark}&\red{\xmark}\\
Android Lab~\cite{android-lab}& \red{\xmark}& \green{\cmark}& \red{\xmark}& \green{\cmark}&\green{\cmark}\\
AndroidWorld~\cite{androidworld}     &\red{\xmark}&\green{\cmark}&\red{\xmark}&\green{\cmark}&\red{\xmark}\\ \hline
\sys (ours) &\green{\cmark}&\green{\cmark}&\green{\cmark}&\green{\cmark}&\green{\cmark}  \\ \hline
\end{tabular}
\end{table*}

To address these limitations, we present \sys, a benchmark that enables faithful, scalable, and modular evaluation of mobile GUI agents. \sys introduces two key innovations:

\textbf{Multi-Branch Static Dataset.} To resolve the dilemma between the rigidity of offline benchmarks and the fragility of online benchmarks, \sys aims to preserve the reproducibility of static datasets while capturing the path diversity of online evaluation. A naive solution would be to exhaustively annotate all possible trajectories for each task. However, this would quickly become intractable due to the combinatorial explosion of paths, making the approach impractical and expensive. Therefore, instead of capturing all possible combinations of trajectories, \sys maintains a single "default" trajectory, but with multiple valid actions---"branches"---at each step. This design effectively captures the multi-path nature of mobile tasks without the overhead of exhaustive path annotation or runtime engineering.


\textbf{Modular Evaluation Framework.} To help researchers navigate the complex design space of mobile GUI Agent, \sys decomposes the agent system into five standardized modules: (i) Screen Parser, (ii) Prompt Styler, (iii) History Generator, (iv) Feedback Generator, and (v) the underlying LFM. This modular architecture enables users to isolate the impact of individual design choices and identify optimal configurations through controlled reconfiguration. 

Our evaluation shows that \sys achieves 94.72\% agreement with human evaluators while revealing that single-path datasets can underestimate agent capabilities by up to 16.09 percentage points (49.9\% relative). Furthermore, our module-level analysis demonstrates that an agent's performance can swing from 4.72\% to 42.72\% solely due to its module configuration, even when the underlying model is held constant---highlighting that the module level evaluation is crucial for building an effective and efficient mobile GUI agents. Finally, we identify the best-performing module combinations across model scales and provide actionable guidelines for building capable and cost-efficient agents.

We summarize our contributions as follows:
\begin{enumerate}
    \item We develop a \textbf{multi-branch static benchmark dataset} that captures multi-path nature of mobile tasks---offering the fidelity of online benchmarks while maintaining scalability and reproducibility of offline benchmarks.

    \item We propose a \textbf{modular benchmark framework} that enables independent evaluation of agent components, facilitating both systematic optimization and fair comparison across mobile GUI agents.

    \item We present comprehensive analysis identifying \textbf{optimal component combinations} and provide insights and actionable guidelines for designing more capable and cost-efficient mobile GUI agents.
\end{enumerate}

\begin{figure*}[t]
	\centering
	\includegraphics[width=0.9\textwidth]{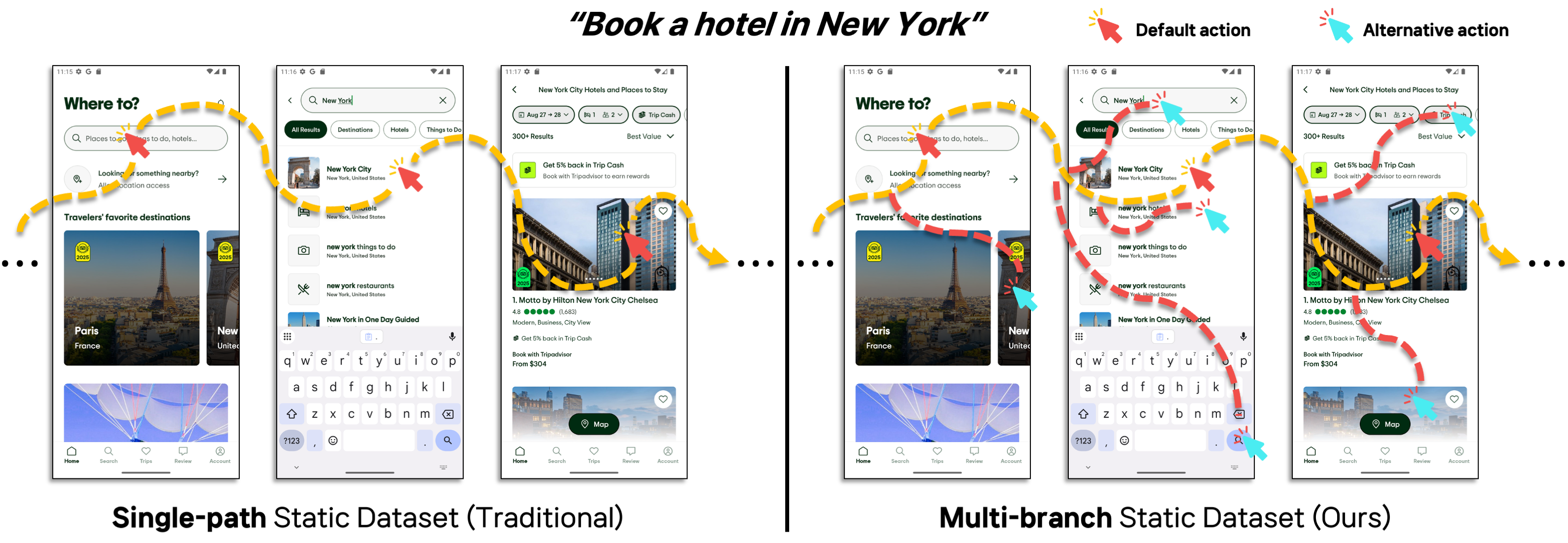}
	\caption{Examples of single-path static dataset and multi-branch static dataset. \textnormal{For the 'Book a hotel in New York' task within the dataset, the left side evaluates task completion based on a single predefined trajectory, whereas the right side accommodates path diversity by accepting all correct actions at each step as valid alternatives.}}
	\label{fig:multi-branch dataset}
\end{figure*}

\section{Background and Motivation}
\label{sec:2}
LFM-powered GUI agents typically adopt a pipeline architecture~\cite{mobile-agent-v2,appagent,android-lab} that translates user instructions into executable UI actions (e.g., clicks, scrolls, text inputs) through multiple stages:
\begin{itemize}[leftmargin=*]
   \item \textbf{Screen parsing:} Converts a raw mobile screenshot into a structured representation interpretable by an LFM, using techniques such as visual markers~\cite{som}, OCR~\cite{omniparser}, or accessibility tree parsing~\cite{android_accessibilityservice_api}.

   \item \textbf{History Generation:} Maintains a coherent memory of prior actions~\cite{generative-Agents, v-droid} to provide the LFM with essential context for subsequent steps.
   
   \item \textbf{Action Inference:} Given the task instruction, parsed screen, and interaction history, predicts the next action~\cite{react,least-to-most,cot} via carefully designed prompts~\cite{mobile-agent-v2, coco-agent, AITZ}, commonly employing techniques such as ReAct~\cite{react} and few-shot learning~\cite{few-shot-learning}.

   \item \textbf{Action Reflection:} Validates the proposed action before execution to detect errors, correct misinterpretations, or recover from mistakes~\cite{self-refine, reflexion, mobile-agent-E}.

\end{itemize}



\subsection{The Need for Modular Evaluation}
While such modular architecture has become standard for mobile GUI agents, existing benchmarks still evaluate agents as monolithic pipeline. This coarse-grained evaluation leads to three critical problems.

\textit{\textbf{Unfair Comparisons.}}
Different implementations of individual modules can lead to significant performance differences, yet existing studies often compare agents without controlling for these variations. As a result, performance improvements are often misattributed to a single factor (e.g., the underlying LFM), overlooking the impact of other components. For instance, an agent using Model A with a Hybrid Modality Parser~\cite{react,least-to-most,cot} may outperform one using Model B with a Text-Only Parser. Black-box benchmarking may attribute the gain solely to Model A, when the screen parsing strategy is the real driver. Without disentangled, module-level evaluation, meaningful comparison is infeasible.

\textit{\textbf{Hidden Performance Factors.}}
Seemingly minor design choices (e.g., how UI elements are represented or how action history is summarized) can significantly impact downstream performance. However, these effects are obscured in end-to-end evaluation, making it difficult to identify critical bottlenecks or optimization opportunities. Moreover, when an agent fails, errors may stem from multiple sources: the screen parser (e.g., missing UI elements), the prompt design (e.g., ambiguous instructions), or the underlying LFM. Without fine-grained, component-level evaluation, designing agentic systems becomes guesswork rather than systematic engineering.

\textit{\textbf{Suboptimal Performance.}}
Optimal module configurations are often context-dependent, varying with factors such as model size, input modality, and application domain~\cite{agentsquare, android-lab}. However, studies often blindly adopt “best practices” across different settings, leading to suboptimal agent performance. In the absence of granular benchmarking, systematically identifying configurations that are well-suited to a given context remains challenging.




\subsection{Benchmarking Mobile GUI Agent}
Benchmarks for Mobile GUI Agents typically falls into one of two categories: offline static benchmarks or online runtime benchmarks. Both have significant tradeoffs.

\textit{\textbf{Offline benchmarks}}~\cite{rico, aitw, motif, metagui, mobile-agent-v2} use static datasets of pre-recorded screenshot sequences, each paired with a single ground-truth action. These actions collectively form a "golden path" for a given task. Such datasets are easy to use and reproduce: the evaluator presents each screenshot to the agent and checks whether the predicted action matches the ground-truth, marking any deviation as failure. This simplicity also makes static datasets easy to scale, as non-expert annotators can readily create new task entries through simple demonstration. However, mobile tasks inherently support multiple valid paths. By annotating only one, static datasets penalize agents that pursue equally valid alternatives. Our analysis shows that mobile tasks have on average \textit{2.95} valid actions at each step, suggesting that current static benchmarks systematically underestimate agent capabilities (\autoref{sec:fidelity}). 


\textit{\textbf{Online benchmarks}}~\cite{mobile-env, mobilegpt,mobile-bench, android-lab, mobileagentbench, androidarena}---such as LLaMA-Touch~\cite{llamatouch} and AndroidWrold~\cite{androidworld}---run agents in live mobile environments and evaluate whether they reach specific \textit{app state checkpoints}, rather than enforcing a fixed action sequence. This design naturally supports multiple valid interaction paths, but it comes with several important drawbacks:

\textit{(i) Limited scalability due to engineering overhead:} Each task requires manually designed checkpoint logic to detect intermediate or final success states. Implementing these checkpoints often demands intimate knowledge of the app logic, specialized instrumentation, or integration with external libraries, limiting scalability of the dataset.
\textit{(ii) Limited reproducibility due to environmental instability:} Live environments are inherently fragile to dynamic content (e.g., ads, pop-ups), cached user data, or app updates that can alter app behavior between runs. As a result, experiments are difficult to reproduce across agents or over time, undermining the reliability of their results. Consequently, most online benchmarks are limited to a small set of open-source or tightly controlled apps---further limiting their coverage.

Taken together, existing benchmarks present a clear dilemma. Offline benchmarks are scalable and reproducible, but their single-path annotation either misrepresents agent performance. Conversely, Online benchmarks capture realistic execution diversity but suffer from engineering complexity and poor reproducibility. This motivates the need for a new evaluation approach that is simultaneously multi-path-aware, scalable, and reproducible.

\section{\sys}
We present \sys, a multi-path aware, modular offline benchmark framework designed to support faithful, scalable, and fine-grained evaluation of mobile GUI agents. Unlike existing benchmarks that treat agents as monolithic systems and rely on fragile single-path datasets or complex runtime environments, \sys explicitly \textbf{decouples the agent architecture} and introduces a novel \textbf{multi-branch evaluation dataset} that captures path diversity without sacrificing reproducibility or scalability. Specifically, \sys is designed around four key goals:
\begin{itemize}[leftmargin=*]
\item \textbf{Fidelity: }
Accurately reflect real-world agent performance by recognizing \textit{multiple valid actions} at each step.

\item \textbf{Scalability: }
Facilitate scalable dataset creation by enabling non-expert annotators to \textit{easily contribute} new tasks and instructions.

\item \textbf{Reproducibility: }
Provide a stable, deterministic evaluation environment that ensures results are consistent and \textit{fairly comparable} across agents, configurations, and time.

\item \textbf{Modularity:}
Support the systematic analysis of Mobile GUI Agents, allowing evaluators to isolate the impact of individual components through controlled \textit{module-level configurations}.
\end{itemize}

\subsection{Multi-Branch Static Dataset}
\label{sec:3.2}
Accurately evaluating Mobile GUI Agents requires capturing the inherent path diversity of mobile tasks. A straightforward way to capture this diversity would be to exhaustively annotate every possible trajectory for each task. However, this quickly becomes intractable, as it requires recording an exponential number of state–action sequences. This limitation is a primary reason why many benchmarks rely on costly and fragile online runtime evaluation.

The key innovation of \sys is its \textit{multi-branch static dataset}, which enables high-fidelity, multi-path-aware evaluation even in offline settings. Instead of recording all possible paths, our dataset maintains a \textit{single default trajectory} while annotating multiple valid \textit{alternative actions}---i.e., ``branches''---at each step (see \autoref{fig:multi-branch dataset}). The core insight is that an agent’s success fundamentally depends on its ability to select valid actions at each decision point; choosing a valid action at every step is effectively equivalent to following a correct path. By focusing on the diversity of valid \textit{actions} rather than paths, \sys preserves the reproducibility and scalability of static offline evaluation while emulating the flexibility of real-world multi-path execution.

\begin{figure*}[t]
    \centering
    \includegraphics[width=0.95\textwidth]{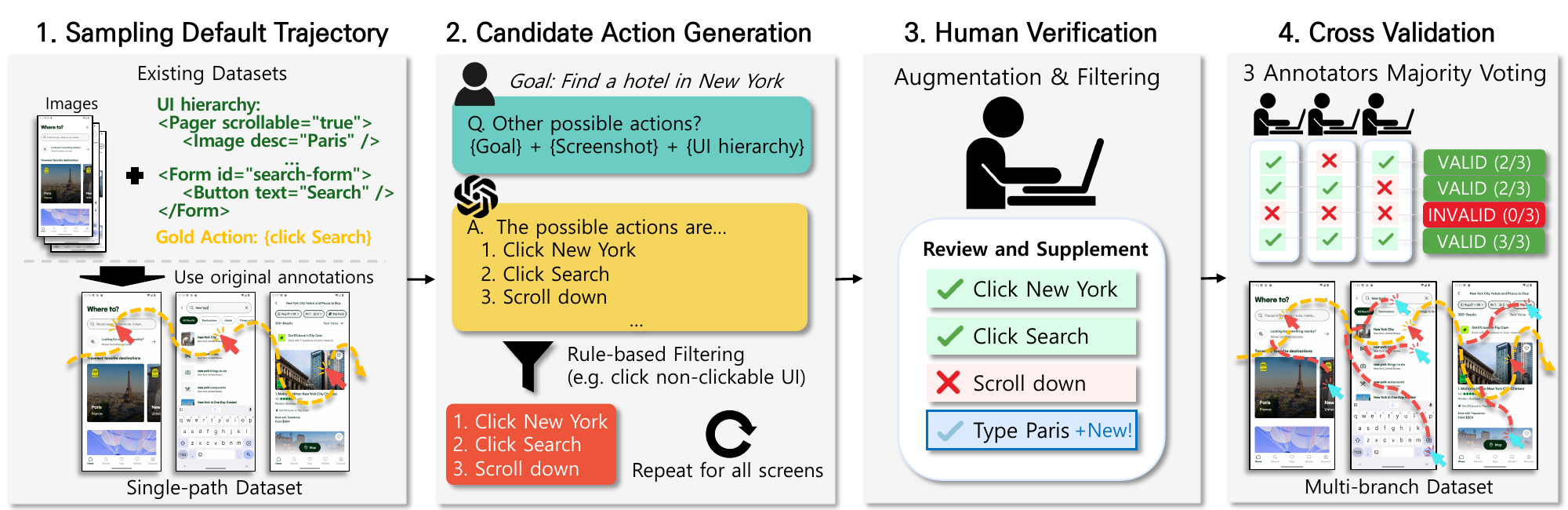}
    \caption{Dataset construction pipeline}
\end{figure*}

\subsubsection{Running the Dataset.}
The evaluation procedure of multi-branch dataset is nearly identical to that of a traditional single-path dataset. At each step, the agent is presented with a screenshot and prompted to predict the next action. If the predicted action matches \textit{any} of the annotated valid actions---including both the default and alternative actions---the step is marked correct and the evaluation proceeds to the next screen in the default trajectory.

A key design choice concerns how interaction history is handled. Since the agent may select any valid action at a given step, but evaluation must advance along the fixed default trajectory, we update the agent's history as if it had taken the default action. This keeps the evaluation anchored to a fixed trajectory while still rewarding alternative valid choices. The approach effectively reduces task completion to a greedy, step-wise decision process, but in doing so guarantees reproducibility and full compatibility with existing offline evaluation practices. Importantly, this simplification does not compromise evaluation quality: our empirical study (\autoref{sec:fidelity}) shows that this design achieves near-perfect agreement with human judgment (94.72\%), indicating that step-wise action validity is a strong proxy for end-to-end task success.

\subsubsection{Dataset Construction.}
We constructed the multi-branch dataset with 27 non-expert annotators. Through several rounds of pilot studies, we identified the following key observations that shaped our annotation workflow:
(1) Annotators find it difficult to choose a single default action when multiple plausible options exists.
(2) Annotators are more effective at identifying additional valid actions when starting from an initial candidate set, rather than labeling everything from scratch.
(3) Annotators excel at filtering out incorrect or invalid actions from a candidate set.
Guided by these observations, we designed a four-stage annotation pipeline.

\noindent \textbf{Stage 1---Sampling Tasks and Default Trajectory.}
Rather than constructing the dataset from scratch, we sampled tasks and screenshots from existing single-path datasets---LlamaTouch, MobileGPT, Meta-GUI, and AndroidWorld---and adopted their original annotations as the default trajectory. This sidesteps the need for annotators to arbitrarily select a default action and provides a strong foundation for multi-branch augmentation \textit{(observation 1)}.

\noindent \textbf{Stage 2---LLM-Based Candidate Action Generation.}
To provide annotators with an initial candidate set, we leveraged state-of-the-art LLMs (GPT-o3 and Gemini 2.5-Pro) to generate an initial set of candidate actions. The models were prompted to enumerate all actions that could reasonably advance the agent toward the task goal, giving annotators a concrete starting point \textit{(observation 2).}

\noindent \textbf{Stage 3---Human Augmentation and Filtering.}
Annotators then reviewed the LLM-generated candidates, supplementing any missing valid actions and removing incorrect or redundant ones \textit{(observation 3).}

\noindent \textbf{Stage 4---Cross-Validation.}
To improve annotation quality and reduce individual bias, each step was cross-validated by three independent annotators. The final set of valid actions was determined by majority vote.

This hybrid human--LLM pipeline enabled efficient scaling of dataset construction while preserving high annotation quality and consistency.

\begin{table}[]
\caption{\sys multi-branch dataset statistics.}
\vspace{-0.3cm}
\small
\begin{tabular}{cccccc}
\hline
\textbf{\begin{tabular}[c]{@{}c@{}}Sampled\\ Dataset\end{tabular}} &
  \textbf{\# App} &
  \textbf{\# Task} &
  \textbf{\begin{tabular}[c]{@{}c@{}}Avg.\\ \# Steps\end{tabular}} &
  \textbf{\begin{tabular}[c]{@{}c@{}}Avg.\\ \# UIs\end{tabular}} &
  \textbf{\begin{tabular}[c]{@{}c@{}}Avg.\\ \# Actions\end{tabular}} \\ \hline
\textbf{LlamaTouch}   & {27} & {163} & {5.6}  & {36.34} & {2.76} \\
\textbf{MobileGPT}    & 7  & {73}  & {5.75} & {34.56} & {2.85} \\
\textbf{Meta-GUI}     & {10} & {167} & {11.02}& {41.51} & {3.38} \\
\textbf{AndroidWorld} & {22} & {105} & {9.43} & {28.23} & {2.38} \\ \hline
\textbf{Total}        & {66} & {508} & {8.21} & {36.52} & {2.95} \\ \hline
\label{tab:statistics}
\end{tabular}
\end{table}

\begin{table}[t]
    \centering
    \footnotesize
    \caption{
        Task difficulty \& complexity categorized based on the number of steps involved and the average number of UIs involved.
    }
    \vspace{-0.3cm}
    \setlength{\tabcolsep}{3pt} 
    \renewcommand{\arraystretch}{1.1} 
    \begin{tabular}{lccc|ccc}
        \hline
        & \multicolumn{3}{c|}{\textbf{Difficulty}} & \multicolumn{3}{c}{\textbf{Complexity}} \\ \cline{2-7}
        \textbf{Source} & \textbf{Easy} & \textbf{Med.} & \textbf{Hard} & \textbf{Simple} & \textbf{Moder.} & \textbf{Complex} \\ \hline
        \textbf{LlamTouch}    & {60}
& {98}
& {5}
& {31}
& {79} & {53}
\\
        \textbf{MobileGPT}    & {26}
& {44}
& {3}
& {15}
& {28} & {30}
\\
        \textbf{Meta-GUI}     & {37}
& {63}
& {67}
& {0}
& {55} & {112}
\\
        \textbf{AndroidWorld\_Static} & {20}
& {56}
& {29}
& {43}
& {46} & {16}
\\ \hline
        \textbf{\sys Dataset}        & {143}
& {261}
& {104}
& {89} & {208} & {211}
\\ \hline
    \end{tabular}
    \begin{tablenotes}[flushleft]
\item \textit{Difficulty:} Easy (≤ 4 steps), Medium (5–11 steps), Hard (≥ 12 steps)
\item \textit{Complexity:} Simple (≤ 25 UIs), Moderate (26–40 UIs), Complex (> 40 UIs)
\end{tablenotes}
\label{tab:difficulty_complexity}
\end{table}

\subsubsection{Statistics.} 
Our final multi-branch dataset comprises 508 unique tasks spanning 66 mobile applications, resulting in a total of 4,173 screenshots and 12,339 annotated actions. A detailed breakdown of dataset statistics is presented in \autoref{tab:statistics}. 

A key highlight is that each step has, on average, 2.95 valid actions (Avg. Actions), indicating a high degree of path diversity in mobile tasks. This implies that the single-path datasets we sampled from have up to a 66\% chance of mis-penalizing a valid agent action, substantially \textit{underestimating} true agent capabilities.

\autoref{tab:statistics} also reports the average number of UI elements per screen (Avg. \# UIs) as a measure of screen complexity. Notably, AndroidWorld\_Static\footnote{AndroidWorld\_Static is a custom static dataset we created using snapshots from the online benchmark AndroidWorld~\cite{androidworld}} exhibits significantly lower complexity---22.7\% fewer UI elements than the average of the other three datasets. This reflects a known limitation of online benchmarks: the high engineering cost of checkpoint instrumentation pushes them toward simplified open-source applications, which are not reliable proxies for real-world environments.

Furthermore, we observe a positive correlation between screen complexity and valid actions per step (see~\autoref{appendix:A}). Screens with more UI elements tend to offer more valid actions, yielding higher path diversity. This further underscores the importance of multi-path evaluation, especially for real-world apps with rich, complex interfaces.

Finally, to enable fine-grained performance analysis across varying task difficulty and complexity, \autoref{tab:difficulty_complexity} categorizes tasks into three difficulty levels and three complexity levels based on the number of steps involved and the average number of UIs involved in each task.

\subsection{Modular Benchmark Architecture}

\label{sec:3.1}
\begin{figure*}[t]
    \centering
    \includegraphics[width=0.8\textwidth]{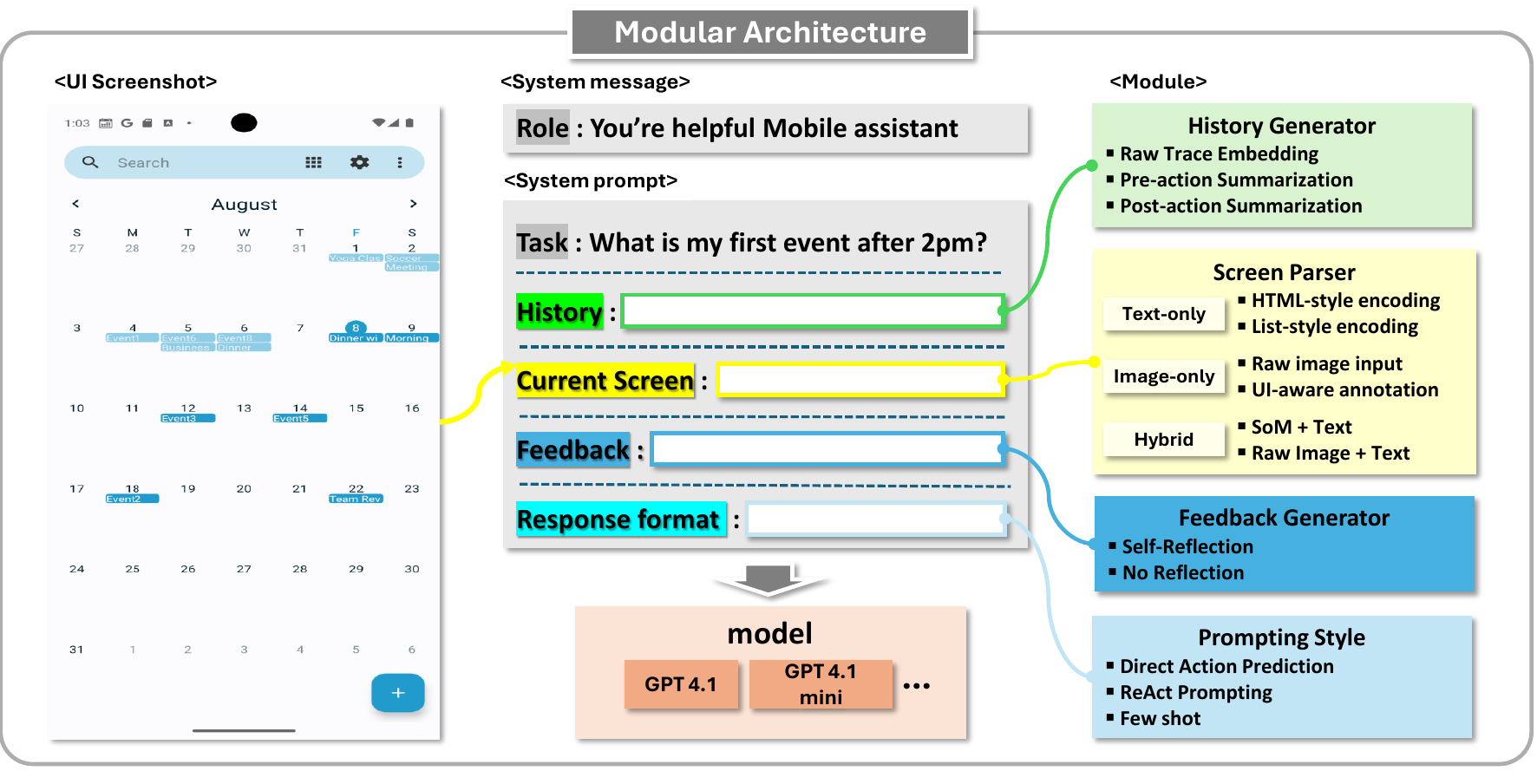}
    \caption{Modular architecture of Mobile GUI Agents. \textnormal{All modules are fully independent and can be flexibly configured into various combinations.}}
    \label{fig:Architecture}
\end{figure*}

Another key contribution of \sys is its support for controlled, component-level evaluation of Mobile GUI Agents. Rather than treating agents as monolithic black boxes, the \sys framework decomposes them into standardized modules: a backbone LFM and four preprocessing modules that construct its input prompt (\autoref{fig:Architecture}). This modularity allows researchers to systematically reconfigure design choices, isolating the impact of each component on the agent's ability to understand and interact with the UI. \sys also seamlessly integrates with custom modules and end-to-end agents.

\subsubsection{Screen Parser (UI Representation)}
The Screen Parser module transforms the raw mobile interface into a structured format that the LFM can interpret. Because how an agent "sees" the UI fundamentally dictates its interaction capabilities, \sys supports three diverse modalities (see \autoref{appendix:B}):

\textbf{Text-only Modality.} 
Text-based parsers leverage the Android Accessibility (a11y) tree~\cite{android_accessibilityservice_api} to extract a hierarchical structure of UI elements, capturing attributes (e.g., text, descriptions) and interactive properties (e.g., clickable). Because raw trees are often overly verbose, \sys implements two refined representations: \textit{HTML-style encoding}~\cite{mobilegpt, enabling_conversational}, which maps UI elements to HTML tags to preserve spatial and structural relationships (e.g., layout containers); and \textit{List-style encoding}~\cite{androidworld, autodroid}, which flattens the hierarchy into a concise list of purely interactable UI elements by discarding non-graphical background nodes.

\textbf{Image-only Modality.} 
When underlying UI metadata is inaccessible (e.g., iOS apps, web views), agents must rely exclusively on visual pixels. We implement \textit{Raw image input}~\cite{uitars, appagent}, which directly feeds screenshots to vision-language models, and \textit{UI-aware visual annotation}~\cite{omniparser, screen-parsing}. The latter employs object detection and OCR to explicitly draw bounding boxes around graphical UI elements and extract on-screen text, providing the model with critical pseudo-structural guidance.

\textbf{Hybrid Modality.} 
Hybrid approaches merge textual and visual representations to strengthen UI understanding. A prominent technique is \textit{Set-of-Marks (SoM)}~\cite{som}, which utilizes UI coordinates from the a11y tree to overlay numbered visual markers directly onto the screenshot. This creates an explicit spatial link between the graphical UI elements and their corresponding textual descriptions, enabling more precise visual grounding for multimodal LFMs.

Alternatively, simpler hybrid strategies concatenate raw screenshots with textual encodings without explicit spatial grounding between the two modalities.
\subsubsection{History Generator (Interaction Memory)}
Agents require prior interaction context to accurately predict subsequent actions. Since concatenating full histories is computationally expensive and exceeds LFM context limits, \sys injects step-wise summaries into the prompt using three techniques:

\textbf{Raw Trace.} 
Directly embeds the agent's previous raw output. While computationally efficient, it lacks the semantic richness needed to convey the underlying interaction intent.

\textbf{Pre-action Summarization.} 
Generates a natural language description of the expected outcome \textit{before} or \textit{in-parallel} with action execution. This provides explicit, human-readable intent with minimal latency overhead.

\textbf{Post-action Summarization.} 
Creates a summary \textit{after} observing the resulting UI state. Although waiting for the UI to update introduces latency, it yields a more accurate, visually grounded reflection of the actual state transition.

\subsubsection{Prompting Style (Reasoning Strategy)}
Prompting fundamentally dictates the agent's decision-making workflow. Because the prompt engineering design space is unbounded, \sys focuses on evaluating three representative response formats:

\textbf{Direct Action Prediction.} 
Requires the LFM to directly output primitive UI actions (e.g., click, input) and parameters. While highly efficient, it obscures the agent’s internal decision-making process and often struggles with complex tasks requiring deep reasoning.

\textbf{ReAct Prompting.} 
Following the Reasoning + Acting paradigm~\cite{react,mobilegpt,android-lab}, the LFM must articulate intermediate thoughts before acting. This chain-of-thought approach~\cite{cot} forces the model to analyze visual and contextual cues, leading to more informed decisions.

\textbf{Few-shot Learning.} 
Augments the prompt with curated examples of interpreting screens and selecting actions~\cite{mobilegpt, few-shot-learning}. Exposing the model to concrete input–output pairs helps it internalize task structures and reduces interaction ambiguity.

\subsubsection{Reflection (Feedback Generator): }
Modern agents increasingly use feedback loops to verify outputs before acting~\cite{self-refine, vsa, reflexion}. Mimicking human error-recovery, several Mobile GUI Agents~\cite{verisafeagent,mobile-agent-v2} now adopt self-reflection. \sys evaluates this by integrating an optional reflection module: if a proposed action misaligns with the user's goal, a feedback generator provides real-time critique, prompting the agent to iteratively refine its prediction prior to execution.

\begin{table*}[t]
\small
\centering
\caption{Task completion rate (TSR) of m3a mobile GUI Agent measured under four different evaluation methodologies. \newline All four methodologies used the same task instruction set.}
\renewcommand{\arraystretch}{1.2}
\setlength{\tabcolsep}{5.5pt}
\begin{tabular}{l|ccc|ccc|cc}
\hline
\multirow{2}{*}{\textbf{Method}} &
  \multicolumn{3}{c|}{\textbf{Task Difficulty}} &
  \multicolumn{3}{c|}{\textbf{Task Complexity}} &
  \multicolumn{2}{c}{\textbf{Overall}} \\
 & \textbf{Easy} & \textbf{Medium} & \textbf{Hard} & \textbf{Simple} & \textbf{Moderate} & \textbf{Complex} & \textbf{TSR} & \textbf{Fidelity}$^{1}$ \\ \hline
\textbf{\# of Tasks}&{25}&{56}&{29}&{43}&{36}&{16}&\multicolumn{2}{c}{{105}}\\ \hline

\textbf{AndroidWorld (Online)}&
{30.00\%}&{34.82\%}&{22.98\%}&{44.57\%}&{23.18\%}&{14.58\%}&{30.63\%}&{94.90\%}\\

\textbf{AndWorld\_Static (Single-Path)}&
{23.33\%}&{20.83\%}&{0\%}&{20.93\%}&{6.5\%}&{29.16\%}&{16.18\%}&{50.15\%}\\

\textbf{\sys (Multi-branch)}&
{33.33\%}&{41.66\%}&{19.54\%}&{41.86\%}&{21.73\%}&{47.91\%}&{33.97\%}&{94.72\%}\\ \hline

\textbf{Human Evaluator (Online)}&
{33.88\%}&{31.5\%}&{24.52\%}&{46.51\%}&{24.39\%}&{17.36\%}&{32.27\%}&---\\ \hline
\end{tabular}

\begin{tablenotes}
\item $^{1}$The metric \textit{Fidelity} indicates how closely each methodology’s results align with human evaluations.
\end{tablenotes}

\label{tab:fidelity}
\end{table*}

\section{Evaluation}

To understand the effectiveness and practical impact of \sys, we conduct a comprehensive evaluation across diverse set of Mobile GUI Agents. Specifically, our evaluation address the following research questions:
\begin{enumerate}[label=\textbf{RQ\arabic*.}]
\item How accurately does \sys's multi-branch dataset assess Mobile \textit{GUI Agent capabilities} compared to existing benchmarks? (\autoref{sec:fidelity})
\item How do individual modules contribute to overall performance, and what is the \textit{best-performing configuration} across different underlying models? (\autoref{sec:4.2})
\item What are the trade-offs between accuracy and cost (e.g., latency, token usage), and which configurations offer the \textit{most cost-efficient} solutions? (\autoref{sec:4.3})
\item What are the characteristics and implications of using \textit{specialized models} (e.g., reasoning models and fine-tuned sLLMs) for Mobile GUI Agents? (\autoref{sec:4.4})
\end{enumerate}

\subsection{Benchmark Faithfulness}
\label{sec:fidelity}
We evaluate \sys along two dimensions of practical importance: the effort required to construct the benchmark (scalability) and how faithfully it reflects real-world agent performance (fidelity). We compare four methodologies: \texttt{AndroidWorld} (online), \texttt{AndroidWorld\_Static} (single-path offline), \sys (multi-branch offline), and human evaluation, all on the same task set from AndroidWorld. As the test agent, we use m3a~\cite{androidworld}, a lightweight but representative Mobile GUI Agent powered by GPT-4.1. Reported results are averaged over three runs. Ground truth was established by three human judges who manually validated full agent execution traces (inter-annotator agreement: 93.01\%).

\textit{\textbf{Annotation and Engineering Effort.}}
We first examine the construction cost of each benchmark. For online evaluation, \texttt{AndroidWorld} requires manual checkpoint engineering: implementing runtime checkpoints for its 116 tasks demands 17,458 lines of code ($\sim$150 LoC/task) across 92 files, each requiring deep knowledge of app internals and software engineering expertise. For offline benchmarks, the dominant bottleneck is manual page navigation to record trajectories. Here, \sys requires traversing only 991 pages, whereas a a naive ``Multi-Path'' dataset that explicitly covers all valid trajectories would require an estimated $\sim$6,533 pages, a 6.6
$\times$ increase due to combinatorial explosion. Since page navigation is the primary source of annotation fatigue, this growth makes exhaustive multi-path annotation impractical at scale, especially for real-world apps with rich path diversity.

\textit{\textbf{Fidelity.}}
\autoref{tab:fidelity} reports the Task Success Rate (TSR) of m3a~\cite{android-inthe-zoo} under four different evaluation settings. As expected, \texttt{AndroidWorld} shows the highest alignment with human evaluators, as its runtime checkpoints are explicitly engineered to emulate human judgments. In contrast, its offline counterpart, \texttt{AndroidWorld\_Static} substantially underestimates agent capability---{by 16.09 percentage points (49.9\% relative)}---due to its rigid single-path assumption.

These results underscore the key trade-off between existing paradigms: Online benchmarks provide high fidelity, but at the cost of substantial engineering overhead and time ($\sim$5x slower than offline benchmarks due to live app interaction). Offline benchmarks are fast and scalable, but their fidelity is limited when evaluation is tied to a single reference path.

\sys successfully bridges this gap, achieving 94.72\% agreement with human evaluators---on par with \texttt{AndroidWorld (Online)}, even under offline settings and static dataset. This corresponds to an 88.8\% improvement over the single-path static baseline, validating our core insight: step-wise action validity is a reliable proxy for overall task success.

One notable exception arises in \textit{Complex} tasks, where \sys tends to overestimate performance. This is because the default trajectory in \sys is often the shortest and most efficient valid path, due to natural tendency of human annotators to record the most direct route to task completion. Since \sys always continues evaluation along this default trajectory, an agent that selects a locally valid but more complex branch is effectively brought back to the simpler path at the next step. This increases agents' likelihood of success compared to online settings, where agents are allowed to explore longer and more complex path. A related limitation is that, unlike online benchmarks, \sys cannot evaluate agent recoverability (e.g., navigating to the wrong screen and returning), as it does not permit such deviations. We discuss these limitations and potential extensions in \autoref{sec:discussion}. Nevertheless, \sys's strong overall fidelity, significantly outperforming single-path offline benchmarks, confirms that our multi-branch design provides a simple yet highly effective way to capture real-world path diversity in offline evaluation.


\begin{table*}[t]
\caption{Incremental modular evaluation across different model sizes.}
\centering
\footnotesize
\begin{threeparttable}
\setlength{\tabcolsep}{5pt}
\begin{tabular}{
  l
  l
  S S S c
  S S S c
  S S S c
}
\toprule
\multirow{2}{*}{\textbf{Module}} &
\multirow{2}{*}{\textbf{Technique}} &
\multicolumn{4}{c}{\textbf{GPT-4.1}} &
\multicolumn{4}{c}{\textbf{GPT-4.1 mini}} &
\multicolumn{4}{c}{\textbf{GPT-4.1 nano}} \\
\cmidrule(lr){3-6} \cmidrule(lr){7-10} \cmidrule(lr){11-14}
& & {\textbf{A.Acc}} & {\textbf{Cost$^{1}$}} & {\textbf{TSR}} & {\textbf{Best}} &
     {\textbf{A.Acc}} & {\textbf{Cost$^{1}$}} & {\textbf{TSR}} & {\textbf{Best}} &
     {\textbf{A.Acc}} & {\textbf{Cost$^{1}$}} & {\textbf{TSR}} & {\textbf{Best}} \\
\midrule
\multirow{6}{*}{\makecell[l]{Screen\\Parser}}
 & a11y--HTML          & 73.64 & \multicolumn{1}{c}{0.0338} & 27.17 &  &59.33  & \multicolumn{1}{c}{0.0068} & 5.91 &  & \best{41.94} & \multicolumn{1}{c}{0.0017} & \best{0.39} & \cmark\\
 & a11y--UI List       & 73.04 & \multicolumn{1}{c}{0.0402} & 28.15 &  & 60.03 & \multicolumn{1}{c}{0.0080} & 4.53 &  & {37.09} & \multicolumn{1}{c}{0.0020} & {0} & \\
 & Image--Raw          & 39.39 & \multicolumn{1}{c}{0.0342} & 4.72  &  & 33.49 & \multicolumn{1}{c}{0.0107} & 2.56 &  & 19.15 & \multicolumn{1}{c}{0.0037} & 0.39 &\\
 & Image-Annotation$^{2}$   & 69.21 & \multicolumn{1}{c}{0.0505} & 16.14 &  & 59.65 & \multicolumn{1}{c}{0.0136} & 5.51&  & 28.96 & \multicolumn{1}{c}{0.0045} & 0.39 &\\
 & Hybrid-SoM+a11y$^{3}$    & \best{74.66} & \multicolumn{1}{c}{0.0615} & \best{32.68} & \cmark & 64.51 & \multicolumn{1}{c}{0.0144} & 10.63&  & 34.88 & \multicolumn{1}{c}{0.0050} & 0 &\\
 & Hybrid-Raw+a11y$^{3}$  & {75.52} & \multicolumn{1}{c}{0.0552} & {31.50} & & \best{65.53} & \multicolumn{1}{c}{0.0158} & \best{15.94} & \cmark & 38.47 & {0.0050} & 0.2 & \\
\addlinespace
\multirow{3}{*}{\makecell[l]{History\\Generation}}
 & Raw Trace           & 74.66 & \multicolumn{1}{c}{0.0615} & 32.68 &  & 65.63 & \multicolumn{1}{c}{0.0158} & 15.94 &  & 41.94 & \multicolumn{1}{c}{0.0017} & 0.39 & \\
 & Pre-action summary  & {79.47} & \multicolumn{1}{c}{0.1222} & {41.93} &  & {70.42} & \multicolumn{1}{c}{0.0321} & {19.69} &  & {42.01} & \multicolumn{1}{c}{0.0033} & {0.79} & \\
 
 & Post-action summary & \best{81.06} & \multicolumn{1}{c}{0.1556} & \best{42.72} & \cmark & \best{73.43} & \multicolumn{1}{c}{0.0416} & \best{24.61} & \cmark & \best{43.41} & \multicolumn{1}{c}{0.0041} & \best{1.38} & \cmark \\ 
 
\addlinespace
\multirow{3}{*}{\makecell[l]{Inference\\Style}}
 & Action Only         & \best{81.06} & \multicolumn{1}{c}{0.1556} & \best{42.72} & \cmark & 73.43 & \multicolumn{1}{c}{0.0416} & 24.61 &  & 43.41 & \multicolumn{1}{c}{0.0041} & 1.38 &\\
 & ReAct               & 81.04 & \multicolumn{1}{c}{0.1589} & 40.94 &  & \best{75.93} & \multicolumn{1}{c}{0.0423} & \best{32.68} & \cmark & \best{48.95} & \multicolumn{1}{c}{0.0043} & \best{3.74} & \cmark\\
  & Few Shot               & 80.44 & \multicolumn{1}{c}{0.2408} & 42.32 &  & {71.67} & \multicolumn{1}{c}{0.0587} & {25.59} & & {38.88} & \multicolumn{1}{c}{0.0084} & {0.2} & \\
\addlinespace
\multirow{2}{*}{Reflection}
 & No Reflection       & \best{81.06} & \multicolumn{1}{c}{0.1558} & \best{42.72} & \cmark & \best{75.93} & \multicolumn{1}{c}{0.0423} & \best{32.68} & \cmark & \best{48.95} & \multicolumn{1}{c}{0.0043} & \best{3.74} & \cmark \\
 & Self Reflection     & 79.93 & \multicolumn{1}{c}{0.2333} & 38.19 &  & 75.02 & \multicolumn{1}{c}{0.0618} & 26.77 &  & 46.98 & \multicolumn{1}{c}{0.0100} & 2.17 & \\
\bottomrule
\end{tabular}
\begin{tablenotes}[flushleft]
\item   
\textit{Metrics.} A.Acc = Action Accuracy; TSR = Task Success Rate; Cost = \$ per task; “Best” marks the within-module best technique for the given model.
\item $^{1}$ GPT-4.1 (Input: \$2.00/1M, Output: \$8.00/1M), GPT-4.1 Mini (Input: \$0.40/1M, Output: \$1.60/1M), GPT-4.1 Nano (Input: \$0.10/1M, Output: \$0.40/1M)
\item $^{2}$ UI Captioning~\cite{omniparser2} + OCR~\cite{easyocr}
\item $^{3}$ For clarity of presentation, we report only the results using the better-performing a11y parsing variant for each hybrid technique. The performance difference between the two a11y parsing methods (HTML vs. UI List) was negligible when combined with the corresponding image-based style.
\end{tablenotes}
\end{threeparttable}
\label{tab:modular_eval}
\end{table*}

\subsection{Modular Evaluation and Take Aways}
\label{sec:4.2}
Having validated the fidelity of our multi-branch dataset, we conduct a comprehensive module-level evaluation to measure the performance impact of each component across three backbone models: GPT-4.1, GPT-4.1 mini, and GPT-4.1 nano.

To keep the combinatorial space of module configurations tractable, we adopted an incremental tuning strategy: beginning from the simplest available techniques for each module (Screen Parser: \textit{a11y-HTML}, History: \textit{Raw Trace}, Inference: \textit{Action Only}, Reflection: \textit{No Reflection}), we optimized one module at a time in the order shown in \autoref{tab:modular_eval}, fixing each to its best variant before proceeding the next.

The results are summarized in \autoref{tab:modular_eval}. Under their best configurations, GPT-4.1 and GPT-4.1 mini achieve TSRs of 42.72\% and 32.68\%, respectively. GPT-4.1 nano performs poorly across all configurations, suggesting that effective GUI interaction requires a minimum level of model capability; we omit its detailed breakdown for brevity. Notably, the same optimal configurations achieve only 27.36\% and 23.03\% on a traditional single-path dataset, underestimating performance by 15.36 and 9.55 percentage points, respectively. This level of gap is comparable to the performance difference between GPT-4.1 and GPT-4.1 mini, reconfirming that single-path evaluation systematically distorts the evaluation results.

Below, we share several key design implications uncovered by this evaluation:

\noindent\textbf{No ``One-Size-Fits-All'' Solution.}
A consistent finding across our experiments is that no single configuration dominates across models. For instance, \textit{ReAct} prompting significantly boosted GPT-4.1 mini's TSR from 24.61\% to 32.68\%, but slightly hurt GPT-4.1, which performed best with the simplest \textit{Action Only} style. In Screen Parsing, adding visual input improved GPT-4.1 mini by 2.7$\times$, but yielded only a 16\% gain for GPT-4.1. In History Generation, \textit{Post-action Summarization} substantially benefited smaller models (mini and nano), while offering little benefit to GPT-4.1. The same pattern appears across model families as well: in our Qwen3 study, Qwen3-VL-8B performed best with \textit{Few-Shot} prompting and showed no difference between pre- and post-action summaries, unlike the similarly sized GPT-4.1 nano. Taken together, these results show that optimal configurations are model-specific and cannot be inferred without empirical testing, underscoring the need for modular benchmarking rather than blindly adopting transferred heuristics.

\noindent\textbf{The Pitfalls of Outdated Heuristics.}
Several techniques commonly assumed as standard practice provided little benefit or even hurt performance. Most notably, \textit{Set-of-Mark (SoM)} prompting, widely adopted to improve visual grounding, showed no meaningful gains and even underperformed the simpler \textit{Raw Image+a11y} parser on smaller models. Likewise, \textit{ReAct} prompting was suboptimal for GPT-4.1, underperforming the simpler \textit{Action Only} style despite its additional cost and latency. These findings reflect a broader dynamic: as foundation models rapidly evolve, techniques validated on earlier generations can quickly lose their edge or become counterproductive. Fine-grained modular evaluation is essential for detecting these shifts.

\noindent\textbf{Naive Self-Reflection Often Backfires.}
Our results also caution against naive implementation of self-reflection. While reflection mechanisms are designed to help agents recover from errors, it may introduce a high risk of false alarms. In our evaluation, self-reflection consistently degraded performance across all models. For GPT-4.1, only 52.94\% of the actions flagged as errors were genuinely erroneous. While the agent corrected 44\% of true errors (80 actions), false alarms caused 127 initially correct actions to be incorrectly revised---a net negative. More advanced reflection pipelines~\cite{cove, ever, verisafeagent} or stronger reflection models may alleviate this issue, but our results show that feedback mechanisms should not be assumed beneficial by default and must be carefully validated before deployment.

\begin{figure}[hbt!]
    \centering
    \includegraphics[width=\linewidth, height=5cm, keepaspectratio]{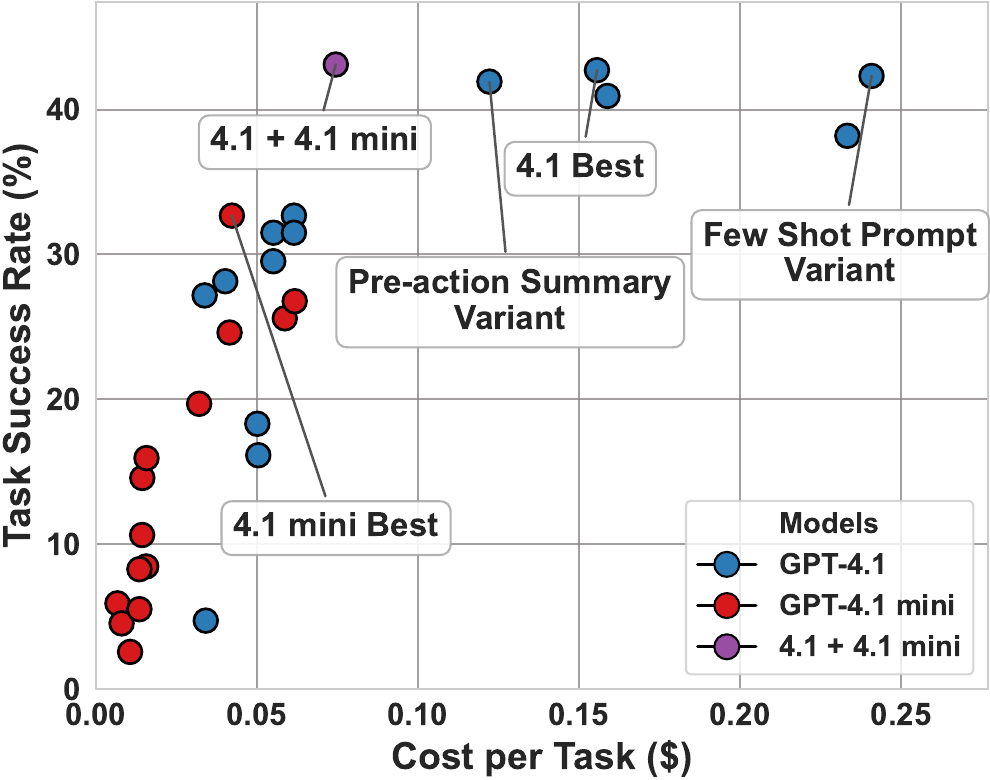}
    \caption{Cost efficiency of different module combinations}
    \label{fig:cost_efficiency}
    \vspace{-0.5cm}
\end{figure}

\subsection{Cost \& Latency Analysis}
\label{sec:4.3}
While task success rate (TSR) is a critical measure of agent performance, real-world viability often hinges on the balance between accuracy and the practical costs of deployment. In this section, we analyze the cost efficiency and latency of various module configurations to assess their real-world efficiency. In particular, we identify representative trade-offs between techniques that deliver strong performance gains and those that impose substantial overhead for only marginal benefit.

\textbf{Cost Efficiency.}
\autoref{fig:cost_efficiency} visualizes the cost–performance distribution of all evaluated module combinations. The results demonstrate that higher cost does not necessarily translate to higher TSR, and the "best-performing" configuration is not always the most practical one. For instance, within GPT-4.1 configurations, The \textit{Post-Action Summary} and \textit{Few-Shot Prompting} variants achieves the highest TSR, making it the best \textit{`performing'} configuration. However, the \textit{Pre-Action Summary} variant achieves comparable TSR, but comes at nearly half the cost, making it the most cost-efficient choice in practice.

A further optimization can be done through hybrid model assignment, where we assign different LFMs to each module. In particular, we found that keeping GPT-4.1 for the core Action Inference module while offloading History Summarization to the cheaper GPT-4.1 Mini preserves performance close to the best GPT-4.1 setup while reducing cost by 52\%.

\begin{figure}[t]
    \centering
    \includegraphics[width=\linewidth]{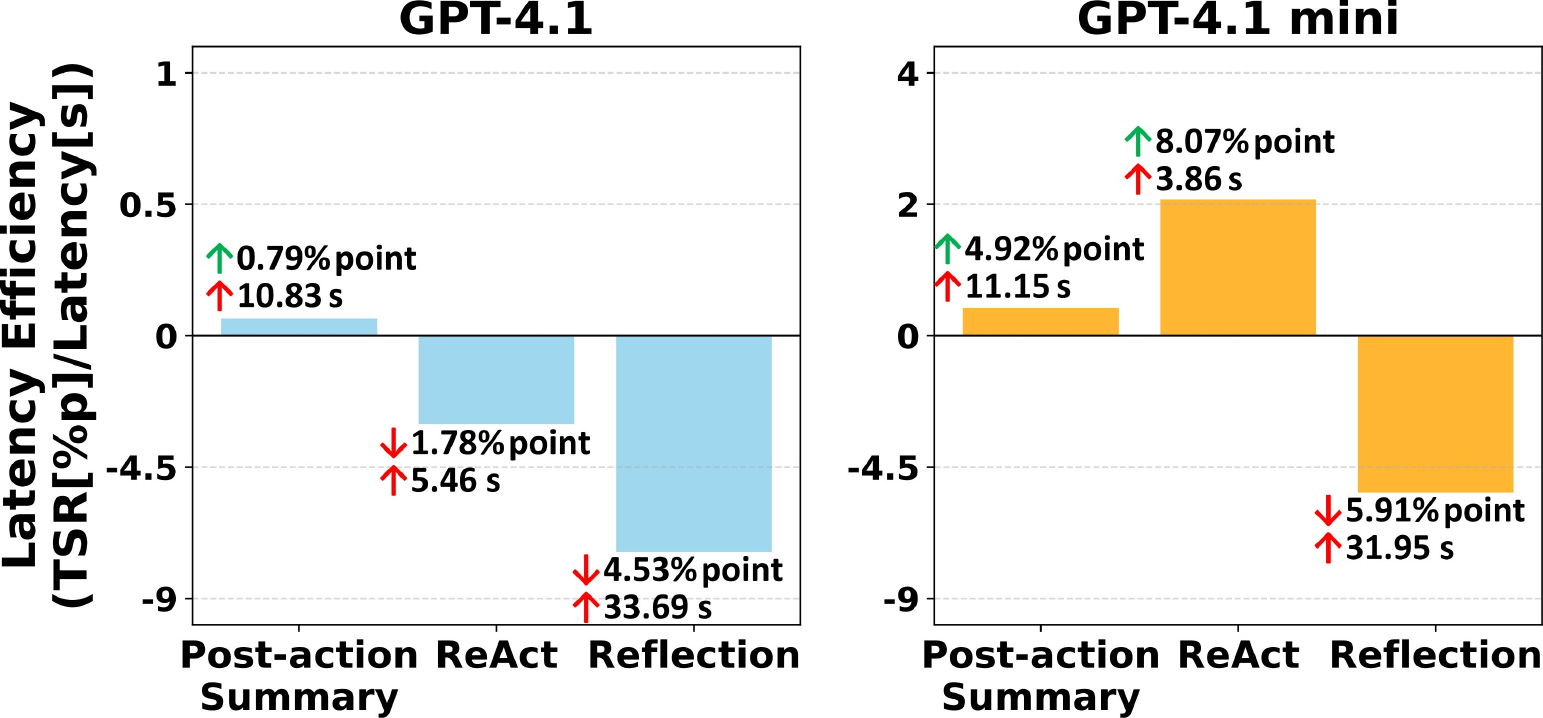}
    \caption{Efficiency of latency incurring techniques.}
\label{fig:latency_efficiency}
\vspace{-0.1cm}
\end{figure}

\noindent\textbf{Latency Analysis.}
\autoref{fig:latency_efficiency} reports \textit{latency efficiency}\footnote{Latency estimates were computed using average throughput (tokens/s) and time-to-first-token (TTFT) reported by OpenRouter over 2025-11-04 to 2025-11-17: GPT-4.1 (throughput: 60.7 tokens/s, TTFT: 0.839s) and GPT-4.1 Mini (throughput: 71.5 tokens/s, TTFT: 0.713s)}---the TSR return per unit of added latency---where positive values indicate positive TSR return and negative values indicate negative TSR return (i.e., degrades TSR despite increased latency). Since LFM's inference latency is dominated by output token count, techniques that affect only the input (e.g., Screen Parsers, Few-Shot prompting) are excluded. \textit{Pre-action Summarization} is also excluded, as it runs in parallel with action execution and added no measurable latency in our experiments.

The results show that most techniques yield weak or even negative TSR returns. The only exception is \textit{ReAct} prompting on GPT-4.1 Mini, which boosts TSR by 8.07\% point at only 3.86s latency cost. For GPT-4.1, \textit{Post-action Summarization} is the only technique with a positive return, but its latency overhead is substantial. Given that \textit{Pre-action Summarization} achieves comparable TSR gains at negligible latency cost, \textit{Pre-action Summarization} variant remains the most efficient choice for GPT-4.1.

\subsection{Discussion}
\label{sec:4.4}

\textbf{\textit{Fine-tuned Models.}}
A number of prior works have proposed small, fine-tuned language models for efficient Mobile GUI Agents. Since these models are trained with specialized inference pipelines, they must be evaluated end-to-end without module separation.

\autoref{tab:fine-tuned} compares the performance of three state-of-the-art GUI agents fine-tuned on Qwen 2.5-VL 7B. Similar to GPT-4.1 Nano, these models initially exhibit very low Overall TSR despite fine-tuning. However, a granular error analysis reveals that failures are disproportionately concentrated in two specific actions: \texttt{Open App} and \texttt{Finish}---the actions that initiate and terminate a task.

To quantify this effect, we compute the TSR excluding these two steps (“w/o Open \& Finish”). All small models, including the Qwen2.5-VL base model, show dramatic TSR increases, some even surpassing GPT-4.1 Mini. Notably, this gain was more pronounced in fine-tuned models than in the base Qwen 2.5-VL. Their action-level accuracy also rises more sharply, indicating that the disproportional error distribution is further amplified in Fine-tuned models.

A likely explanation is \textit{data imbalance} during finetuning. Unlike intermediate actions (e.g., \texttt{click}, \texttt{input}), which occur multiple times per task, \texttt{Open App} and \texttt{Finish} appear exactly once per task, giving models far fewer learning opportunities. 

Taken together, these findings not only highlight an inherent limitation of small fine-tuned models but also provide actionable guidance for future development. Since opening an app and determining the end of the task is the primary bottleneck, upsampling fine-tuning data for \texttt{Open} and \texttt{Finish} actions could resolve this bottleneck and unlock significant performance gains for small, efficient GUI models.

\begin{table}[]
\caption{Performance of small, fine-tuned LFMs}
\vspace{-0.4cm}
\label{tab:fine-tuned}
\small
\begin{tabular}{l|cc|cc}
\hline
\multirow{3}{*}{\textbf{\begin{tabular}[c]{@{}l@{}}Models\end{tabular}}} &
  \multicolumn{2}{c|}{\textbf{Overall}} &
  \multicolumn{2}{c}{\textbf{w/o Open \& Finish}} \\ \cline{2-5} 
 &
  \multirow{2}{*}{\textbf{A.Acc}} &
  \multirow{2}{*}{\textbf{TSR}} &
  \multirow{2}{*}{\textbf{A.Acc}} &
  \multirow{2}{*}{\textbf{TSR}} \\
                        &         &                 &         &                  \\ \hline
\textbf{Qwen 2.5 VL-7B} & 57.38\%    & 3.35\%            & 59.17\%    & 29.33\%             \\
\textbf{GUI OWL 7B~\cite{mobile-agent-v3}}     & 61.30\% & 0.00\%          & 69.02\% & 36.22\%          \\
\textbf{UI Genie 7B~\cite{ui_genie}}    & 64.15\% & 1.58\%          & 72.13\% & \textbf{38.98\%} \\
\textbf{UITars 1.5 7B~\cite{uitars}}  & 56.63\% & \textbf{7.68\%} & 62.65\% & 32.09\%          \\ \hline
\end{tabular}
\end{table}

\begin{table}[]
\caption{Performance across GPT-5.1's reasoning efforts}
\vspace{-0.4cm}
\label{tab:reasoning}
\small
\begin{tabular}{l|cccc}
\hline
\textbf{\begin{tabular}[c]{@{}l@{}}Reasoning\\ Effort\end{tabular}} &
  \textbf{TSR} &
  \textbf{\begin{tabular}[c]{@{}c@{}}Reasoning \\ Tokens / Task\end{tabular}} &
  \textbf{\begin{tabular}[c]{@{}c@{}}Cost \\ per Task\end{tabular}} &
  \textbf{\begin{tabular}[c]{@{}c@{}}Latency\\ per Task\end{tabular}} \\ \hline
\textbf{\begin{tabular}[c]{@{}l@{}}None\end{tabular}} &
  40.55\% &
  -- &
  \$0.065 &
  43.00s \\ \hline
\textbf{Low} &
  \begin{tabular}[c]{@{}c@{}}39.17\%\\ \footnotesize{($\downarrow$3.40\%)}\end{tabular} &
  0.66K &
  \begin{tabular}[c]{@{}c@{}}\$0.071\\ \footnotesize{($\uparrow$9.23\%)}\end{tabular} &
  \begin{tabular}[c]{@{}c@{}}55.78s\\ \footnotesize{($\uparrow$29.72\%)}\end{tabular} \\ \hline
\textbf{Medium} &
  \begin{tabular}[c]{@{}c@{}}42.13\%\\ \footnotesize{($\uparrow$3.90\%)}\end{tabular} &
  1.51K &
  \begin{tabular}[c]{@{}c@{}}\$0.080\\ \footnotesize{($\uparrow$23.08\%)}\end{tabular} &
  \begin{tabular}[c]{@{}c@{}}75.00s\\ \footnotesize{($\uparrow$74.41\%)}\end{tabular} \\ \hline
\textbf{High} &
  \begin{tabular}[c]{@{}c@{}}44.29\%\\ \footnotesize{($\uparrow$9.22\%)}\end{tabular} &
  4.09K &
  \begin{tabular}[c]{@{}c@{}}\$0.11\\ \footnotesize{($\uparrow$69.23\%)}\end{tabular} &
  \begin{tabular}[c]{@{}c@{}}134.32s\\ \footnotesize{($\uparrow$212.37\%)}\end{tabular} \\ \hline
\end{tabular}
\end{table}

\textbf{\textit{Reasoning Models}}
Recent advances in the test-time compute paradigm~\cite{ttc} have enabled models to solve complex problems through extended reasoning. To examine the impact of test-time compute in the context of Mobile GUI Agents, we evaluated GPT-5.1~\cite{gpt5-1}---a state-of-the-art reasoning model---under varying reasoning efforts. For the base agent, we used the best-performing modular configuration of GPT-4.1 identified in our earlier evaluation (\autoref{sec:4.2}).

\autoref{tab:reasoning} presents the performance of GPT-5.1 across different levels of reasoning effort. Unlike domains such as math or coding, where scaling test-time compute yields dramatic gains, Mobile GUI Agent performance improves only modestly with additional reasoning. The absolute TSR improvement is limited to 3--4 percentage points (maximum 9.22\% relative gain). However, this marginal gain comes at a disproportionate expense: a \textit{69.23\%} increase in monetary cost and a \textit{212.37\%} increase in latency.

These findings suggest that GUI interaction relies less on deliberate, step-by-step reasoning (System-2 thinking) and more on intuitive visual grounding and pattern recognition (System-1 thinking). This hypothesis is further supported by our earlier finding that ReAct prompting---designed to strengthen explicit reasoning---failed to improve GPT-4.1's performance. Together, these results imply that future advancements for Mobile GUI Agents may stem not from increasing reasoning depth, but from scaling domain-specific training to better internalize GUI interaction patterns.

\textbf{\textit{Image-only Environment (iOS, Web app).}}
Some application environments (e.g., iOS apps, web applications) restrict access to textual UI representations (a11y). In such cases, agents are forced to rely solely on image-based parsing. To address this, we report modular evaluation results under image-only conditions (see \autoref{appendix:C}). Interestingly, the optimal module configuration remained identical to the standard setting (\autoref{tab:modular_eval}), with the exception of the parser itself: both GPT-4.1 and GPT-4.1 Mini performed best with \textit{Raw Image + UI Captioning}. Under this configuration, they achieve TSRs of 26.97\% and 20.86\%, respectively. This significant performance drop compared to the standard setting highlights the critical role of textual structural information in mobile GUI interaction.

{\footnotesize
\typeout{\footnotesize: \the\fontdimen6\font}
}
\section{Related Works}


\textbf{Multi-path Offline Benchmarks.}
Evaluating long-horizon tasks with multi-path diversity remains a long-standing challenge in offline benchmarking. While various approaches have been explored to address this, scalable solutions remain elusive. Notably, Mobile-Bench-v2~\cite{mobile-bench-v2} leverages a state–action transition graph~\cite{mobilevlm} to support multi-path evaluation. However, constructing such graphs requires exhaustive annotation and incurs substantial overhead and cost, limiting practical scalability. Another line of work employs simulated environments that replicate the front-end behavior of real web pages~\cite{realbench}. Yet, accurately replicating the complex logic of real-world applications demands significant engineering effort, making this approach difficult to scale across diverse domains.



\textbf{Modular Evaluation of Agents.}
Recently, motivated by the increasingly fragmented nature of modern agentic architectures, few works in the Web agent domain have introduced frameworks for module- or model-level evaluation. 
Notably, AgentSquare\cite{agentsquare} decomposes agents into Planning, Reasoning, Tool Use, and Memory modules and quantifies each module’s contribution within a controlled online environment. BrowserGym\cite{browsergym} provides a unified platform that allows researchers to easily evaluate diverse agentic techniques and switch between models.

Inspired by this line of works, we present, to the best of our knowledge, the first modular benchmarking framework tailored to Mobile GUI Agents. We formally decompose the end-to-end agent pipeline into key functional modules and design our benchmark to support isolated, systematic evaluation of each component. Furthermore, to make these evaluations more effective and reliable, we introduce the concept of a \textit{multi-branch static dataset}---a novel benchmarking paradigm that captures path diversity without requiring complex online environments.

\section{Limitation \& Future Work}
\label{sec:discussion}

\textit{\textbf{Inherent limitation of Offline Benchmark.}}
While \sys enables high-fidelity benchmarking in an offline setting, several inherent limitations of offline benchmarking remain. First, it does not capture an agent’s ability to recover from intermediate mistakes. Unlike online evaluation, where agents are permitted to make incorrect decisions (e.g., navigating to the wrong screen) as long as they eventually complete the task within a step budget, \sys applies a stricter criterion in which any single incorrect action results in task failure. Similarly, static datasets cannot capture how agents respond to unexpected events such as app crashes, pop-ups, or ads. A promising direction for future work is to build datasets that include trajectories with occasional errors, enabling evaluation of error recovery.

Second, the default trajectories in the \sys multi-branch dataset are subject to human annotator bias, particularly a preference for shorter, more efficient paths. As a result, agents are primarily evaluated along these simplified trajectories, which may overestimate their capabilities and fail to reflect performance on more complex or less intuitive paths. This issue is especially pronounced when extreme shortcuts exist. While the bias can be mitigated by retaining additional reference trajectories when annotators show large discrepancies in path length, it remains a fundamental limitation of static, trajectory-driven evaluation.

\textit{\textbf{Non-standard Auxiliary Modules.}}
While \sys supports modular evaluation of Mobile GUI Agents, it does not cover all possible modules used in prior work. Several studies have introduced specialized components such as planning modules~\cite{mapagent, agents2, mobile-agent-E, splanner}, memory modules~\cite{mobilegpt, autodroid, appagent}, or inference pipelines~\cite{autodroid2, v-droid} that deviate from standard action-driven agent workflows. These auxiliary modules can substantially improve performance, but they often rely on task-specific assumptions or specialized designs that make them difficult to integrate into a standardized evaluation pipeline. Nevertheless, \sys is designed to be easily extensible. Custom modules can be added with minimal engineering effort, enabling future work to evaluate such specialized components within a unified and reproducible benchmarking setup.

\textit{\textbf{Multi-app task.}} While various benchmarks\cite{gui-odyssey, mvisubench, androidworld, spabench} support multi-app tasks, the current dataset focuses on single-application scenarios. The benchmark is highly scalable, enabling new multi-app datasets—easily created by non-experts using a simple annotation tool—to integrate seamlessly without modifications to the core system.

\section{Conclusion}
This work introduced \sys, the first modular, multi-path-aware, offline benchmarking framework for Mobile GUI Agents. \sys enables fine-grained, component-level analysis while combining the fidelity of online benchmarks with the scalability and reproducibility of offline benchmarks. We hope \sys serves as a foundation for more rigorous, reproducible, and insightful evaluation in the development of future GUI agents.

\bibliographystyle{ACM-Reference-Format}
\bibliography{references}
\appendix
\onecolumn        
\section{Mobibench Benchmark Details}
\label{appendix:A}

\vspace{0.1cm}
\subsection{Benchmark Selection}

To ensure that our multi-branch dataset is built upon a reliable and representative foundation, we selected and augmented four existing mobile-agent benchmarks—LlamaTouch, MobileGPT, Meta-GUI, and AndroidWorld. This selection was guided by the following practical criteria.

\vspace{0.3em}

\textbf{(1) XML Availability.}  
Each dataset provides structured UI metadata (XML or equivalent hierarchical representations), enabling consistent parsing of UI elements, bounding boxes, widget types, and interactable components. This property is essential for reliably reconstructing screen states during branch annotation.

\vspace{0.3em}

\textbf{(2) Action Space Completeness.}  
These benchmarks define clear and complete action vocabularies (e.g., click, input, scroll), ensuring compatibility with our unified action schema. This consistency also improves the reliability of LLM-based candidate action generation across tasks.

\vspace{0.3em}

\textbf{(3) Popularity and Coverage.}  
The selected datasets span a diverse range of application domains—including productivity, social, navigation, shopping, and media—capturing common interaction patterns observed in real-world mobile applications. This ensures that the sampled tasks are representative of realistic user behavior, and importantly, these benchmarks have been widely adopted in recent mobile-agent research, providing further empirical validation of their reliability and utility.

\vspace{0.3em}

\begin{figure}[ht!]
    \centering
    \includegraphics[width=0.7\linewidth]{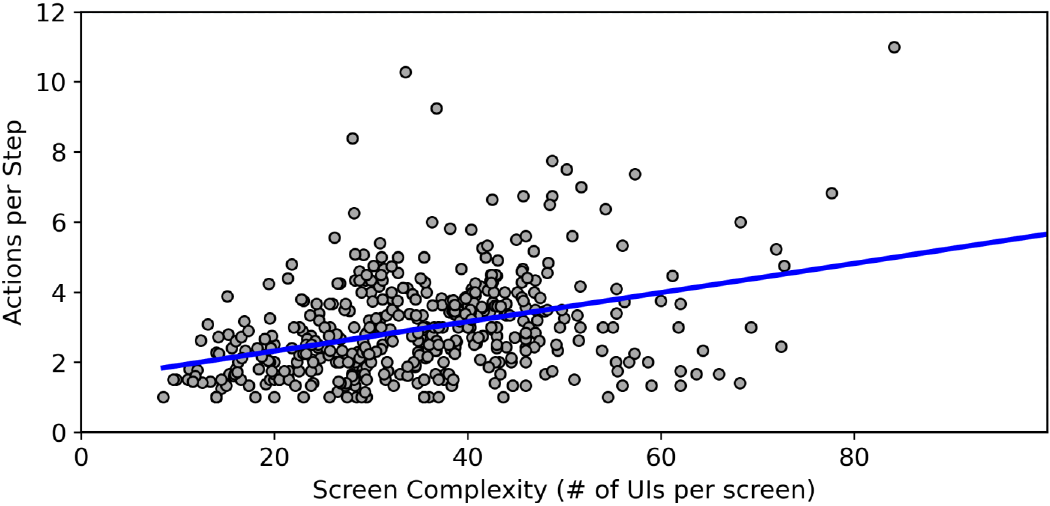}
    \caption{Correlation between screen complexity and path diversity}
    \label{fig:correlation}
\end{figure}

By satisfying these criteria, the selected benchmarks not only offer strong representativeness and data quality but can also be easily expanded through our multi-branch augmentation pipeline. Furthermore, the augmented dataset exhibits a clear relationship between UI complexity and path diversity (\autoref{fig:correlation}), demonstrating that our dataset selection and augmentation process preserves key structural properties of real-world mobile interactions.

\vspace{0.2cm}
\subsection{Task Selection}
\label{appendix:A.2}
To ensure the realism, representativeness, and structural completeness of the multi-branch dataset, we selected benchmark tasks—and their corresponding applications—according to the following criteria.

\vspace{0.3em}

\textbf{(1) App Real-world Usefulness \& Representativeness.}
The applications associated with the selected tasks belong to high-frequency usage categories in everyday mobile interaction (e.g., messaging, social media, email, shopping, media, and productivity) and include widely adopted mobile apps frequently used in prior research.  
This ensures that the augmented tasks reflect realistic user behaviors and that the resulting multi-branch evaluation generalizes well to real-world mobile GUI interaction settings.

\textbf{(2) Completeness of UI Metadata.}
Each task was required to provide stable XML/Accessibility Tree metadata for all steps, ensuring that essential UI information—--such as UI elements, text content, and bounding boxes—was consistently available.  
This guarantees the consistency and reproducibility of LLM-based candidate action generation and valid-action verification.

\textbf{(3) Capture Real World Complexity.}
To reflect real-world complexity of mobile tasks, we included tasks that have multiple alternative UI paths to achieve the same intent. Representative examples include:
\begin{itemize}[leftmargin=1.2em, itemsep=2pt, topsep=2pt]
    \item Pressing a search button vs. selecting the top search bar,
    \item Entering the same screen through different menu routes,
    \item Using the right-side options button vs. a left-side navigation button.
\end{itemize}

\vspace{0.3em}

This criterion ensures that each step contains multiple valid actions, capturing the inherent path diversity of mobile interfaces.

\vspace{0.3em}

\textbf{(4) Cross-benchmark App Sampling.}
Some applications appeared across multiple benchmark datasets---such as LlamaTouch, MobileGPT, Meta-GUI, and AndroidWorld.  
In such cases, we cross-sampled tasks from the same application across these datasets.  
For instance, applications like \emph{Calendar} appeared in multiple benchmarks, enabling integrated sampling. This approach:
\begin{itemize}[leftmargin=1.2em, itemsep=2pt, topsep=2pt]
    \item mitigates dependence on the design biases of a single benchmark,
    \item incorporates broader UI variations at the app level,
    \item enhances dataset representativeness and diversity.
\end{itemize}

\vspace{0.3em}

This strategy was applied only when an app appeared in multiple benchmarks, and was not universally applicable to all apps.

\vspace{0.2cm}
\subsection{Prompts}
\label{appendix:prompt_design}
Below, we share a prompt used in the hybrid LLM-assisted workflow to construct the multi-branch dataset. It enables the model to enumerate all potentially valid actions for a given screen, which are then verified and refined by human annotators.

\begin{tcolorbox}[promptstyle={Candidate Action Generation Prompt},
                  breakable,
                  fontupper=\footnotesize, before upper={\setlength{\parskip}{0.5em}\setlength{\parindent}{0pt}},]

You are a mobile UI validation assistant. 

Your goal is to find all possible UI actions that can accomplish given goals on mobile applications. Based on the provided goal, provided UI screenshot, current UI elements, and actions that can be performed on the UI, you must choose which action on the specific element is appropriate to achieve the goal.

Available actions :

 - Click: Tap on buttons, links, menu items, or interactive elements.\\
 - Input: Type text into text fields, search boxes, or input areas.\\
 - Navigate back: Go back to the previous screen\\
 - Swipe: Scroll up or down to find more relevant content or UI elements.\\
 - Finish: Complete the task when the goal has been accomplished.

The current goal : \{Goal\}

Current UI elements:\\
\{UI representations\}

Action rules:\\
  - The \texttt{element\_id} must be one of the IDs from the ``Current UI elements'' list.\\
  - Actions on an element and its element ID must clearly match.\\
  - Action types include \texttt{click}, \texttt{input}, \texttt{swipe}, and \texttt{navigate\_back}.\\
  - An \texttt{input} action can accomplish the goal only if the element is an
        \texttt{InputField} or a \texttt{TextField} whose text contains ``search''.
        In that case, consider a \texttt{click} action on the same element along with
        the \texttt{input} action.\\
  - For an \texttt{input} action, specify \texttt{text\_to\_input}.\\
  - For \texttt{text\_to\_input}, extract appropriate phrases or words from the goal
        that can be entered in the input field.\\
  - Extract all possible actions, not just a single path to the goal. Consider
        various possibilities and optional paths.\\
  - Make sure to answer according to the JSON array format below.

\vspace{0.3em}
Example response:\\[2pt]
\texttt{[}\{``action\_type'': ``click'', ``element\_id'': 14\},
 \{``action\_type'': ``input'', ``element\_id'': 16, ``text\_to\_input'': ``winter''\},
 \{``action\_type'': ``swipe''\}\texttt{]}[4pt]

response:
\end{tcolorbox}

\newpage

\section{Agent Design}
\label{appendix:B}
In this section, we provide additional description of the basic structure and design of the base Mobile GUI Agent we used for the modular evaluation in \autoref{sec:4.2}.

\begin{figure*}[ht]
    \centering
    \includegraphics[width=1\textwidth]{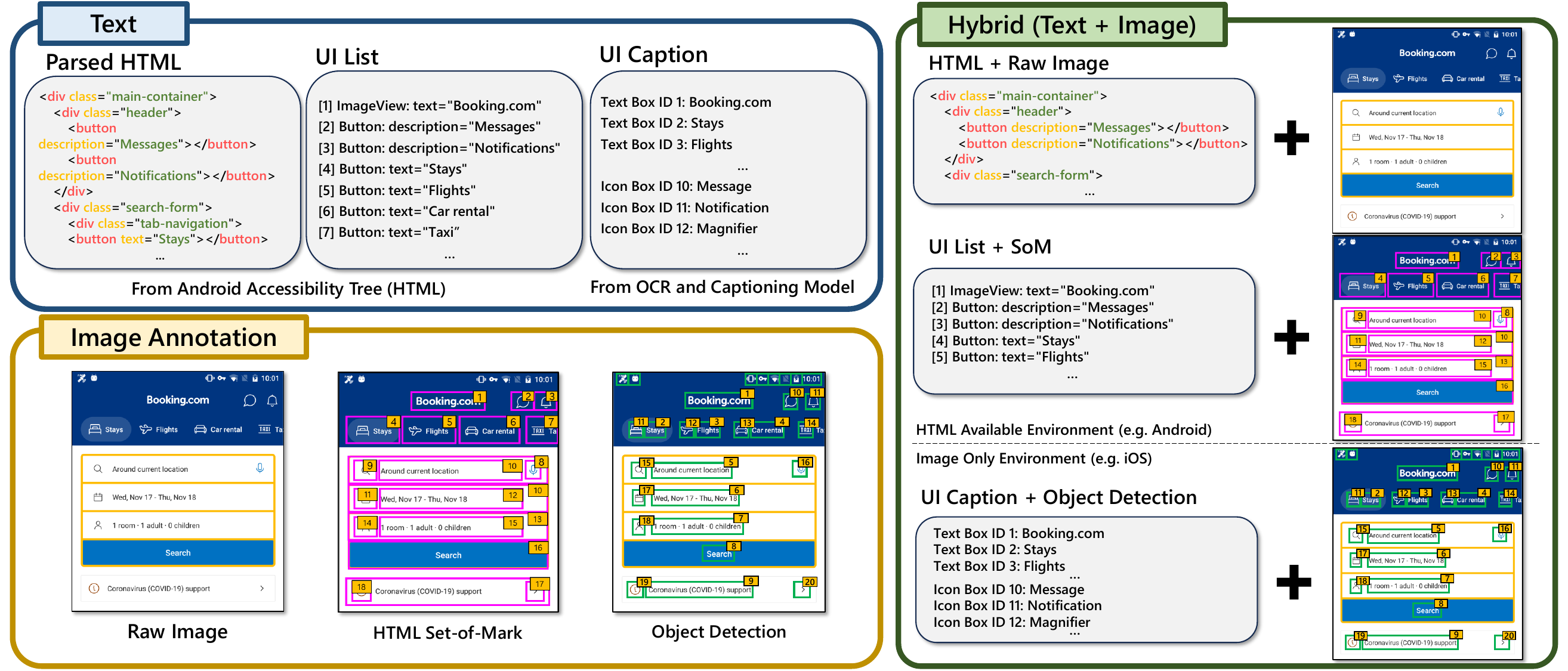}
    \caption{Example techniques for screen parsing.}
    \vspace{-0.4cm}
    \label{fig:parser}
\end{figure*}

\vspace{-0.1cm}
\subsection{Input}
\subsubsection{A11y Tree.}
Android’s Accessibility framework enables extraction of the on-screen UI hierarchy by producing XML dumps that encode view structure, attributes, and interaction affordances. We use this mechanism to collect UI snapshots offline and construct a dataset containing serialized UI trees for each interaction state. During evaluation, the agent operates solely on these pre-extracted XML observations, allowing the benchmark to faithfully approximate real deployment conditions without requiring live traversal of the device interface. This representation functions analogously to an HTML or DOM tree, providing a structured and device-independent view of the mobile UI.

\vspace{0.3em}

\subsubsection{Image.}
In a real Android execution environment, agents obtain visual observations by capturing screenshots of the current screen. Our benchmark adopts the same observation modality by providing these screenshots as input, ensuring that the offline setting closely matches real deployment conditions.

\vspace{0.2cm}
\subsection{Action Space}
The system supports a set of structured actions for interacting with the user interface
and completing tasks. Each action is represented in JSON format and corresponds to a
specific UI-level operation.

\vspace{0.3em}

\begin{itemize}[leftmargin=1.2em, itemsep=2pt, topsep=2pt]
    \item \textbf{Click}: 
    Performs a tap gesture on a UI element identified by its index, triggering the
    corresponding interface transition or interaction.

    \item \textbf{Input}: 
    Inserts a specified text string into the target text field, emulating user typing.

    \item \textbf{Scroll}: 
    Executes a directional scrolling gesture to reveal additional on-screen content
    or navigate within scrollable regions.

    \item \textbf{Navigate Back}: 
    Returns the interface to the previous screen or hierarchical level.

    \item \textbf{Open App}: 
    Launches a specified application by name.

    \item \textbf{Finish}: 
    Signals the completion of the task and terminates the interaction sequence.
\end{itemize}

\begin{tcolorbox}[promptstyle={Action Descriptions in Prompt},
                  breakable,
                  fontupper=\footnotesize, before upper={\setlength{\parskip}{0.5em}\setlength{\parindent}{0pt}},]
action example list:

- Click/tap on a UI element (specified by its index) on the screen:
 \{"action type": "click", "index": <target UI index>\}.
  Use only when the element is already visible and unambiguous. The click should immediately trigger the next UI change (open a page, toggle a control, confirm a dialog). identify the correct index from the UI descriptions and Do not forget to cite that index in the JSON.

- Type text:
 \{"action type": "input", "index": <target UI index>, "params": \{"text": "<text input>"\}\}.
Use only when the goal explicitly requires entering text into the focused field (search, login, chat input). Do not combine with manual clicks on the keyboard; describe the exact text you will enter and always wrap it in `params.text`. Do not forget to include the correct index as well.

- Scroll the current screen or list:
 \{"action type": "scroll", "direction": <up, down, left, right>\}.
  Use when the element you need is off-screen or more content needs to be revealed. Choose the direction that moves toward the target (e.g., `down` to reveal lower content). Do not over-scroll; perform one scroll per action.

- Navigate back:
 \{"action type": "navigate back"\}.
  Use when the current screen is a dead end, you opened the wrong page, or the goal requires returning to the previous view. Do not use `click` on on-screen back buttons unless the instruction explicitly calls for that specific UI button.

- Open an app:
 \{"action type": "open app", "params": \{"app": <app name>\}\}.
 Use at the start of a task or whenever you must switch apps. Do not scroll the home screen or tap icons manually to find an app. Always issue open app with the app name and let the system launch it. If the app name is given in the goal, copy it exactly.

- Finish the task:
 \{"action type": "finish", "status": "complete"\}.
  Only use after verifying that all requirements in the goal have been satisfied. Once the task is complete, choose `finish` immediately instead of taking additional exploratory actions.
\end{tcolorbox}

\subsection{Prompts}
Below, we present the prompts used for our agent, adapted from the Mobile GUI Agent m3a~\cite{aitw}. These base prompts were further adjusted depending on the specific technique being evaluated (e.g., ReAct, Pre-action Summary, Few-Shot Prompting).

\begin{tcolorbox}[promptstyle={Action Agent Prompt},
                  breakable,
                  fontupper=\footnotesize, before upper={\setlength{\parskip}{0.5em}\setlength{\parindent}{0pt}},]

You are an agent who can operate an Android phone on behalf of a user. Based on user's goal/request, complete the tasks by performing actions (step by step) on the phone.

At each step, a list of descriptions for most UI elements on the current screen will be given to you if possible (each element can be specified by an index), together with a history of what you have done in previous steps. Based on these pieces of information and the goal, you must choose to perform one of the action in the following list (action description followed by the JSON format) by outputting the action in the correct JSON format.

\{Action Explanation\}

The current user goal/request is: \{Goal\}\\
Here is a history of what you have done so far:\{History\}\\
Here is a representation of UI elements on the current screen:\\
\{UI representation\}

Now output an action from the above list directly in the correct JSON format. Your answer must be only the JSON object representing the action.
Do not include any additional text or explanation in your answer.

example: \{"action type":...\}

Your Answer:
\end{tcolorbox}

\begin{tcolorbox}[promptstyle={History Generation (i.e., Action Summary) Prompt},
                  breakable,
                  fontupper=\footnotesize, before upper={\setlength{\parskip}{0.5em}\setlength{\parindent}{0pt}},]
You are an agent capable of operating an Android phone on behalf of a user.

Based on user's goal/request, you may

\{Action Explanation\}

The (overall) user goal/request is: \{Goal\}'

Now I want you to summarize the latest step.
You will be given the screenshot before you performed the action, the action you chose, and the screenshot after the action was performed. 

Here is the UI representation (description) of the screen before the action (related with the first image):
\{Before Action UI representation\}

On this screen you chose the following action:
\{Selected Action\}

Here is the UI representation (description) of the screen after the action (related with the second image):
\{After Action UI representation\}

By comparing the two screenshots (plus the UI representation) and the action performed, give a brief summary of this step. This summary will be added to action history and used in future action selection, so try to include essential information you think that will be most useful for future action selection like what you intended to do, why, if it worked as expected, if not what might be the reason (be critical, the action/reason might not be correct), what should/should not be done next and so on.

Make sure to keep it short and in one line.\\
Summary of this step: 

\end{tcolorbox}


\begin{tcolorbox}[promptstyle={Self-Reflection Prompt},
                  breakable,
                  fontupper=\footnotesize, before upper={\setlength{\parskip}{0.5em}\setlength{\parindent}{0pt}},]
You are an agent capable of operating an Android phone on behalf of a user.

Based on user's goal/request, you may

\{Action Explanation\}

The (overall) user goal/request is: \{Goal\}'

Here is a history of what you have done so far:
\{History\}

Now I want you to verify whether the next chosen action is correct.
You will be given the screenshot before you performed the action, the action you chose, and the screenshot after the action was performed. 

Here is the UI representation (description) of the screen before the action (related with the first image):
\{Before Action UI representation\}

On this screen you chose the following action:
\{Selected Action\}

Here is the UI representation (description) of the screen after the action (related with the second image):
\{After Action UI representation\}

By comparing the two screenshots (plus the UI representation) and the action performed, validate whether the action is on the correct path towards the user's goal. The action is correct only if it clearly moves one step closer to the goal.

Be extremely strict: if any part of the action looks wrong, risky, or even slightly misaligned (wrong index/bounds/coordinates/text/direction/app), mark it incorrect. The action is correct only if it clearly moves one step closer to the goal.

You must respond in JSON format:

\{correct (boolean): <true or false, indicating if the action is correct>, 
explanation (string): <Your explanation about your decision>,
feedback (string): <feedback on what went wrong and how to fix it if applicable, or "none" if no fixes are needed.>
\}

Verification result: 
\end{tcolorbox}


\section{Supplementary Experiments Details}
\label{appendix:C}
In this section, we present supplementary experiments that further support our main findings and provide deeper insights into the modular design space of MobiBench.

\subsection{Module combination under Image-only environment}

\begin{table*}[ht]
\caption{Modular evaluation under image-only environment.}
\vspace{-0.3cm}  
\footnotesize
\begin{threeparttable}
\setlength{\tabcolsep}{10pt}
\renewcommand{\arraystretch}{0.8} 
\begin{tabular}{
  l
  l
  S S c
  S S c
}
\toprule
\multirow{2}{*}{\textbf{Module}} &
\multirow{2}{*}{\textbf{Technique}} &
\multicolumn{3}{c}{\textbf{GPT-4.1}} &
\multicolumn{3}{c}{\textbf{GPT-4.1 mini}} \\
\cmidrule(lr){3-5} \cmidrule(lr){6-8}
& & {\textbf{A.Acc}} & {\textbf{TSR}} & {\textbf{Best}} &
     {\textbf{A.Acc}} & {\textbf{TSR}} & {\textbf{Best}} \\
\midrule
\multirow{3}{*}{\makecell[l]{Screen Parser}}
 & Image--Raw          & 39.39 & 4.72 &  & 33.49 & 2.56 &  \\
 & Object Detection SoM + UI Captioning   & 69.21 & 16.14 &  & 59.65 & 5.51 &  \\
 & Raw Image + UI Captioning   & \best{70.66} & \best{18.31} & \cmark & \best{62.78} & \best{8.27} & \cmark \\
\addlinespace
\multirow{3}{*}{\makecell[l]{History Generation}}
 & Raw Trace           & 70.66 & 18.31 &  & 62.78 & 8.27 &  \\
 & Pre-action & 74.03 & 25.98 &  & 65.65 & 17.13 &  \\
 & Post-action & \best{74.85} & \best{26.97} & \cmark & \best{69.79} & \best{17.52} & \cmark \\
\addlinespace
\multirow{3}{*}{\makecell[l]{Inference Style}}
 & Action Only         & \best{74.85} & \best{26.97} & \cmark & 69.79 & 17.52 &  \\
 & ReAct               & 76.08 & 25.39 &  & \best{70.24} & \best{20.86} & \cmark \\
 & Few Shot            & 75.26 & 25.79 &  & 62.32 & 20.47 &  \\
\addlinespace
\multirow{2}{*}{Reflection}
 & No Reflection       & \best{74.85} & \best{26.97} & \cmark & \best{70.24} & \best{20.86} & \cmark \\
 & Self Reflection   & 75.36 & 26.77 &  & 70.44 & 20.47 &   \\
\bottomrule
\end{tabular}

\begin{tablenotes}[flushleft]
\item \textit{Metrics.} A.Acc = Action Accuracy; TSR = Task Success Rate; “Best” marks within-module best technique.
\end{tablenotes}

\end{threeparttable}
\vspace{-0.3cm}
\label{tab:ios}
\end{table*}

\subsection{Fine-tunend Models}

\begin{table*}[ht]
\caption{Action and task success accuracy with vs. without the finish action.}
\vspace{-0.3cm}  
\centering
\footnotesize
\begin{threeparttable}
\setlength{\tabcolsep}{4.3pt} 
\renewcommand{\arraystretch}{0.8} 
\begin{tabular}{
  l
  l
  S S S S
  S S S S
  S S S S
}
\toprule

& &
\multicolumn{4}{c}{\textbf{GPT-4.1}} &
\multicolumn{4}{c}{\textbf{GPT-4.1 mini}} &
\multicolumn{4}{c}{\textbf{GPT-4.1 nano}} \\
\cmidrule(lr){3-6} \cmidrule(lr){7-10} \cmidrule(lr){11-14}

\textbf{Module} &
\textbf{Technique} &
\multicolumn{2}{c}{\textbf{Overall}} &
\multicolumn{2}{c}{\textbf{w/o Fin}} &
\multicolumn{2}{c}{\textbf{Overall}} &
\multicolumn{2}{c}{\textbf{w/o Fin}} &
\multicolumn{2}{c}{\textbf{Overall}} &
\multicolumn{2}{c}{\textbf{w/o Fin}} \\
\cmidrule(lr){3-4} \cmidrule(lr){5-6}
\cmidrule(lr){7-8} \cmidrule(lr){9-10}
\cmidrule(lr){11-12} \cmidrule(lr){13-14}

& &
\textbf{A.Acc} & \textbf{TSR} &
\textbf{A.Acc} & \textbf{TSR} &
\textbf{A.Acc} & \textbf{TSR} &
\textbf{A.Acc} & \textbf{TSR} &
\textbf{A.Acc} & \textbf{TSR} &
\textbf{A.Acc} & \textbf{TSR} \\
\midrule


\multirow{6}{*}{\makecell[l]{Screen\\Parser}}
 & a11y--HTML          & 73.64 & 27.17 & 76.69 & \textbf{44.89} & 59.33 & 5.91 & 65.58 & 34.06 & 41.94 & 0.39 & 41.92 & 11.22 \\
 & a11y--UI List       & 73.04 & 28.15 & 75.70 & 42.52 & 60.03 & 4.53 & 66.97 & 34.06 & 37.09 & 0 & 37.08 & 9.06 \\
 & Image--Raw          & 39.39 & 4.72  & 37.88 & 6.29  & 33.49 & 2.56 & 34.48 & 5.71  & 19.15 & 0.39 & 19.14 & 4.33 \\
 & Image-Annotation$^{1}$ & 69.21 & 16.14 & 71.34 & 30.12 & 59.65 & 5.51 & 67.39 & 29.13 & 28.96 & 0.39 & 28.95 & 6.69 \\
 & Hybrid-SoM+a11y$^{2}$  & 74.66 & \textbf{32.68} & 76.62 & 43.90 & 64.51 & 10.63 & 70.57 & \textbf{43.50} & 34.88 & 0 & 34.88 & 3.54 \\
 & Hybrid-Raw+a11y$^{2}$  & 75.52 & 31.50 & 76.94 & 41.54 & 65.53 & \textbf{15.94} & 70.61 & 40.15 & 38.47 & 0.2 & 38.46 & \textbf{12.0} \\
\addlinespace


\multirow{3}{*}{\makecell[l]{History\\Generation}}
 & Raw Trace           & 74.66 & 32.68 & 76.62 & 43.90 & 65.63 & 15.94 & 70.61 & 40.15 & 41.94 & 0.39 & 41.92 & 11.22 \\
 & Pre-action summary  & 79.47 & 41.93 & 79.96 & 45.67 & 70.42 & 19.69 & 74.53 & 39.38 & 42.01 & 0.79 & 41.98 & 10.63 \\
 & Post-action summary & 81.06 & \textbf{42.72} & 82.04 & \textbf{48.03} & 73.43 & \textbf{24.61} & 77.11 & \textbf{40.16} & 43.41 & 1.38 & 43.36 & \textbf{11.22} \\
\addlinespace


\multirow{3}{*}{\makecell[l]{Inference\\Style}}
 & Action Only         & 81.06 & \textbf{42.72} & 82.04 & \textbf{48.03} & 73.43 & 24.61 & 77.11 & 40.16 & 43.41 & 1.38 & 43.36 & 11.22 \\
 & ReAct               & 81.04 & 40.94 & 81.66 & 44.68 & 75.93 & \textbf{32.68} & 78.28 & \textbf{43.11} & 48.95 & 3.74 & 48.74 & \textbf{13.58} \\
 & Few Shot            & 80.44 & 42.32 & 81.28 & 47.04 & 71.67 & 25.59 & 74.66 & 39.57 & 38.88 & 0.2 & 38.85 & 7.09 \\
\addlinespace


\multirow{2}{*}{Reflection}
 & No Reflection       & 81.06 & \textbf{42.72} & 82.04 & \textbf{48.03} & 75.93 & \textbf{32.68} & 78.28 & \textbf{43.11} & 48.95 & 3.74 & 48.74 & \textbf{13.58} \\
 & Self Reflection     & 79.93 & 38.19 & 80.81 & 42.13 & 75.02 & 26.77 & 77.69 & 37.99 & 46.98 & 2.17 & 46.81 & 10.24 \\
\bottomrule

\end{tabular}

\begin{tablenotes}[flushleft]
\item $^{1}$ UI Captioning~\cite{omniparser2} + OCR~\cite{easyocr}, $^{2}$ Better-performing a11y variant.
\item \textit{Metrics.} A.Acc = Action Accuracy; TSR = Task Success Rate.
\end{tablenotes}

\end{threeparttable}
\vspace{-0.3cm}
\label{tab:wofin}
\end{table*}

\clearpage        

\vspace{0.3em}
\section{Dataset App List}
\label{appendix:D}
The following Table 10 summarizes the application's information used in MobiBench. For each app, we list its name, a short description, and the number of tasks in which it appears. In multi-app tasks, multiple applications may be used simultaneously; therefore, the task counts may include duplicates when an app participates in such tasks.

\renewcommand{\arraystretch}{1} 
\begin{longtable}{p{0.26\textwidth} p{0.65\textwidth} r}
\caption{List of AndroidWorld apps and the number of tasks for each.%
\label{tab:androidworld-apps}}\\
\hline
\textbf{App name} & \textbf{Description} & \textbf{\# tasks} \\
\hline
\endfirsthead

\hline
\textbf{App name} & \textbf{Description} & \textbf{\# tasks} \\
\hline
\endhead

Trello & Provides project and task-management tools using boards, lists, and cards to organize work. & 17 \\ \hline
Pinterest & Offers a visual discovery platform that helps users find and save ideas through images. & 16 \\ \hline
Discord & Supports voice, video, and text communication, widely adopted by online communities. & 10 \\ \hline
Instagram & Lets users share photos, stories, and short videos on a large social platform. & 10 \\ \hline
YouTube & Enables users to watch, upload, and share videos with a global audience. & 10 \\ \hline
Reddit & Provides community-based spaces for discussions, news, and content sharing. & 10 \\ \hline
DoorDash & Connects users with nearby restaurants for convenient meal delivery. & 9 \\ \hline
Walmart & Supports shopping with tools for browsing and purchasing products online. & 9 \\ \hline
Yelp & Helps users discover and review local businesses, restaurants, and services. & 8 \\ \hline
Zoom & Enables users to conduct online meetings and virtual communication via video. & 7 \\ \hline
YT Music & Provides music streaming with playlists, songs, and personalized recommendations. & 22 \\ \hline
ESPN & Delivers sports coverage including live scores, news, and major event updates. & 6 \\ \hline
Amazon Shopping & Supports browsing, purchasing, and tracking items on an online marketplace. & 5 \\ \hline
Expedia & Serves as a platform for reserving flights, hotels, car rentals, and vacation packages. & 5 \\ \hline
Snapchat & Lets users share photos and videos that disappear after viewing. & 4 \\ \hline
CNN & Delivers global and local news coverage, articles, and live updates. & 4 \\ \hline
Gmail & Supports sending, receiving, and organizing email messages. & 12 \\ \hline
Coursera & Offers online learning through courses, certificates, and educational programs. & 4 \\ \hline
Spotify & Provides music streaming for songs, playlists, and podcasts. & 3 \\ \hline
BBC & Delivers international articles, videos, and news reports. & 3 \\ \hline
Maps & Provides navigation tools for finding routes, directions, and location information. & 2 \\ \hline
Clock & Offers alarms, timers, and stopwatch functions for daily utility. & 5 \\ \hline
Facebook & Supports social networking for connecting, posting, and interacting with communities. & 2 \\ \hline
Duolingo & Provides language lessons and exercises across various languages. & 2 \\ \hline
Drive & Enables cloud storage for uploading, storing, and sharing files. & 2 \\ \hline
Calculator & Performs mathematical calculations for everyday use. & 1 \\ \hline
Chrome & Allows users to access and explore websites through a web browser. & 3 \\ \hline
Calendar & Manages event scheduling and reminders for personal planning. & 1 \\ \hline
Proton Calendar & Helps users manage events with privacy-focused calendar features. & 89 \\ \hline
Uber & Connects riders with nearby drivers for quick and convenient transportation. & 23 \\ \hline
Weather: Live radar \& widgets & Shows live radar updates and offers home-screen widgets for instant weather checks. & 14 \\ \hline
Booking.com & Provides tools for booking hotels, vacation rentals, and lodging. & 9 \\ \hline
Simple Calendar & Supports event and appointment management for organizing schedules. & 26 \\ \hline
Weather Forecast: Live Weather & Provides real-time weather updates, forecasts, and predictions. & 8 \\ \hline
Google Calendar & Helps users plan events, manage schedules, and set reminders. & 6 \\ \hline
Weather smart-pro android apps & Offers weather predictions and related utility functions. & 4 \\ \hline
Daily Forecast: Weather \& Radar & Delivers daily weather forecasts and radar maps. & 4 \\ \hline
Google\_dialer & Supports phone calls, call log management, and contact handling on Android devices. & 17 \\ \hline
Microsoft\_to-do & Helps users organize personal tasks and to-do lists. & 12 \\ \hline
Telegram & Supports secure chats, group conversations, and media sharing. & 6 \\ \hline
Tripadvisor & Guides users in exploring and reviewing hotels, restaurants, and attractions. & 7 \\ \hline
Twitter & Enables sharing short updates, news, and real-time posts. & 8 \\ \hline
Audio\_recorder & Captures voice or ambient sounds and saves them as audio files. & 1 \\ \hline
Camera & Captures photos or video through the device camera. & 1 \\ \hline
Contacts & Manages saving, editing, and organizing personal contact information. & 2 \\ \hline
Expense pro & Tracks spending by recording expenses, updating entries, and managing budgets. & 9 \\ \hline
Files & Organizes device storage by letting users move, delete, or modify files. & 5 \\ \hline
Joplin & Provides note-taking tools for writing and storing personal notes. & 4 \\ \hline
Markor & Supports text-focused writing, editing, and organization of notes and folders. & 13 \\ \hline
Osmand & Helps users navigate, save routes, and explore locations through mapping tools. & 3 \\ \hline
Retro music player & Plays and manages local audio files and playlists. & 4 \\ \hline
Settings & Configures device options such as Bluetooth, Wi-Fi, display settings, and system preferences. & 11 \\ \hline
Simple draw pro & Enables sketching and saving simple illustrations. & 1 \\ \hline
Simple sms messenger & Supports sending, replying to, and managing SMS messages. & 6 \\ \hline
Tasks & Helps users create to-dos, set deadlines, and track progress. & 6 \\ \hline
Vlc & Plays a wide range of audio and video file formats. & 2 \\ \hline
Gallery & Manages and browses image files stored on the device. & 3 \\ \hline
Broccoli & Stores and organizes cooking recipes with editing and categorization features. & 13 \\ \hline
OpenTracks & Tracks workouts and records statistics such as distance, duration, and movement patterns. & 6 \\ \hline
\end{longtable}
\par\noindent

\end{document}